\documentclass[10pt,twocolumn,letterpaper]{article}

\usepackage[pagenumbers]{cvpr} 

\usepackage{graphicx}
\usepackage{amsmath}
\usepackage{amssymb}
\usepackage{booktabs}

\usepackage{enumitem}
\usepackage{multirow}
\usepackage{bbding}
\usepackage[dvipsnames]{xcolor}
\usepackage{indentfirst}
\usepackage[accsupp]{axessibility} 
\usepackage{bbm}
\usepackage[pagebackref,breaklinks,colorlinks]{hyperref}

\usepackage[capitalize]{cleveref}
\crefname{section}{Sec.}{Secs.}
\Crefname{section}{Section}{Sections}
\Crefname{table}{Table}{Tables}
\crefname{table}{Tab.}{Tabs.}

\def\cvprPaperID{463} 
\def\confName{CVPR}
\def\confYear{2023}

\begin{document}

\title{One-to-Few Label Assignment for End-to-End Dense Detection}

\author{Shuai Li$^{1,2}$,  Minghan Li$^{1}$,  Ruihuang Li$^{1}$,  Chenhang He$^{1,2}$,  Lei Zhang$^{1,2}$\thanks{Corresponding author.} \\
{$^{1}$The Hong Kong Polytechnic University \qquad $^{2}$OPPO Research Institute}\\
{\tt\small \{csche, csrhli, cslzhang\}@comp.polyu.edu.hk, \{lishuai9401,liminghan0330\}@gmail.com}\\
}

\maketitle
\begin{abstract}
    One-to-one (o2o) label assignment plays a key role for transformer based end-to-end detection, and it has been recently introduced in fully convolutional detectors for end-to-end dense detection. However, o2o can degrade the feature learning efficiency due to the limited number of positive samples.  Though extra positive samples are introduced to mitigate this issue in recent DETRs, the computation of self- and cross- attentions in the decoder limits its practical application to dense and fully convolutional detectors. In this work, we propose a simple yet effective one-to-few (o2f) label assignment strategy for end-to-end dense detection. Apart from defining one positive and many negative anchors for each object, we define several soft anchors, which serve as positive and negative samples simultaneously. The positive and negative weights of these soft anchors are dynamically adjusted during training so that they can contribute more to ``representation learning'' in the early training stage, and contribute more to ``duplicated prediction removal'' in the later stage. The detector trained in this way can not only learn a strong feature representation but also perform end-to-end dense detection. Experiments on COCO and CrowdHuman datasets demonstrate the effectiveness of the o2f scheme. Code is available at \url{https://github.com/strongwolf/o2f}.
\end{abstract}

\section{Introduction}
Object detection~\cite{fasterrcnn, ssd, yolo, fcos} is a fundamental computer vision task, aiming to localize and recognize the objects of predefined categories in an image. Owing to the rapid development of deep neural networks (DNN)~\cite{vgg,resnet,densenet,mobilenets,googlenet1,googlenet2,googlenet3}, the detection performance has been significantly improved in the past decade. During the evolution of object detectors, one important trend is to remove the hand-crafted components to achieve end-to-end detection.

One hand-crafted component in object detection is the design of training samples. For decades, anchor boxes have been dominantly used in modern object detectors such as Faster RCNN~\cite{fasterrcnn}, SSD~\cite{ssd} and RetinaNet~\cite{focalloss}. However, the performance of anchor-based detectors is sensitive to the shape and size of anchor boxes. To mitigate this issue, anchor-free~\cite{fcos,foveabox} and query-based~\cite{detr,deformdetr,dynamicdetr,conditionaldetr} detectors have been proposed to replace anchor boxes by anchor points and learnable positional queries, respectively.

Another hand-crafted component is non-maximum suppression (NMS) to remove duplicated predictions. The necessity of NMS comes from the one-to-many (o2m) label assignment~\cite{primesample,ota,atss,paa,gfocal,gfocalv2}, which assigns multiple positive samples to each GT object during the training process. This can result in duplicated predictions in inference and impede the detection performance. Since NMS has hyper-parameters to tune and introduces additional cost, NMS-free end-to-end object detection is highly desired.

\begin{figure}[tbp]
    \centering
    \includegraphics[width=0.4\textwidth]{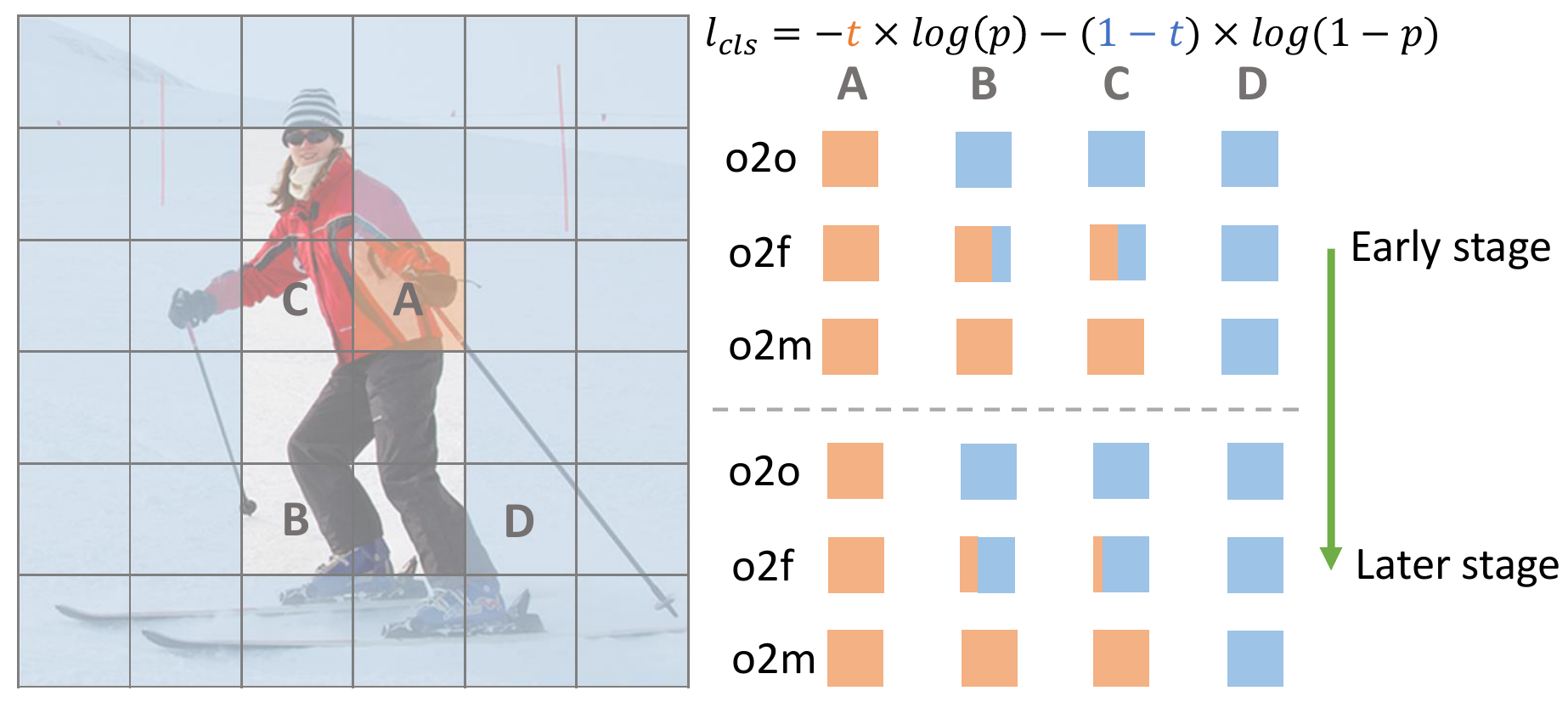}
    \caption{The positive and negative weights of different anchors (A, B, C and D) in the classification loss during early and later training stages. Each anchor has a positive loss weight $t$ (in orange color) and a negative loss weight $1-t$ (in blue color). In our method,  A is a fully positive anchor, D is a fully negative anchor, and B and C are ambiguous anchors. One can see that for o2o and o2m label assignment schemes, the weights for all anchors are fixed during the training process, while for our o2f scheme, the weights for ambiguous anchors are dynamically adjusted.}
    \label{fig_in_intro}
    \vspace{-3mm}
\end{figure}

With a transformer architecture, DETR~\cite{detr} achieves competitive end-to-end detection performance. Subsequent studies~\cite{poto,onenet} find that the one-to-one (o2o) label assignment in DETR plays a key role for its success. Consequently, the o2o strategy has been introduced in fully convolutional network (FCN) based dense detectors for lightweight end-to-end detection. However, o2o can impede the training efficiency due to the limited number of positive samples. This issue becomes severe in dense detectors, which usually have more than 10k anchors in an image. What’s more, two semantically similar anchors can be adversely defined as positive and negative anchors, respectively. Such a ‘label conflicts’ problem further decreases the discrimination of feature representation. As a result, the performance of end-to-end dense detectors still lags behind the ones with NMS. Recent studies~\cite{dndetr,groupdetr,hybriddetr} on DETR try to overcome this shortcoming of o2o scheme by introducing independent query groups to increase the number of positive samples. The independency between different query groups is ensured by the self-attention computed in the decoder, which is however infeasible for FCN-based detectors.

In this paper, we aim to develop an efficient FCN-based dense detector, which is NMS-free yet end-to-end trainable. We observe that it is inappropriate to set the ambiguous anchors that are semantically similar to the positive sample as fully negative ones in o2o. Instead, they can be used to compute both positive and negative losses during training, without influencing the end-to-end capacity if the loss weights are carefully designed. Based on the above observation, we propose to assign dynamic soft classification labels for those ambiguous anchors. As shown in Fig.~\ref{fig_in_intro}, unlike o2o which sets an ambiguous anchor (anchor B or C) as a fully negative sample, we label each ambiguous anchor as partially positive and partially negative. The degrees of positive and negative labels are adaptively adjusted during training to keep a good balance between ‘representation learning’ and ‘duplicated prediction removal’. In particular, we begin with a large positive degree and a small negative degree in the early training stage so that the network can learn  the feature representation ability more efficiently, while in the later training stage, we gradually increase the negative degrees of ambiguous anchors to supervise the network learning to remove duplicated predictions. 
We name our method as a one-to-few (o2f) label assignment since one object can have a few soft anchors. We instantiate the o2f LA into dense detector FCOS, and our experiments on COCO~\cite{coco} and CrowHuman~\cite{crowdhuman} demonstrate that it achieves on-par or even better performance than the detectors with NMS.

\section{Related Work}
The past decade has witnessed tremendous progress in object detection with the rapid development of deep learning techniques~\cite{swin,transformer1,transformer2,transformer3,vgg,resnet}. Modern object detectors can be roughly categorized into two types: convolutional neural networks (CNNs) based detectors~\cite{fasterrcnn,focalloss,yolo,yolo3,yolo4,yolo6,yolo7,ssd,focalloss,cascadercnn,fcos} and transformer based detectors~\cite{detr,conditionaldetr,deformdetr,dabdetr,anchordetr,efficientdetr,dynamicdetr}. 

\subsection{CNN-based Object Detectors}
CNN-based detectors can be further divided into two-stage detectors and one-stage detectors. Two-stage detectors~\cite{fasterrcnn,cascadercnn} generate region proposals in the first stage and refine the locations and predict the categories of these proposals in the second stage, while one-stage detectors~\cite{ssd,focalloss} directly predict the categories and location offsets of dense anchors on convolutional feature maps. The early detectors mostly utilize pre-defined anchors as training samples.  The hyper-parameters of anchor shapes and sizes have to be carefully tuned since the suitable settings vary across different datasets. To overcome this issue, anchor-free detectors~\cite{fcos, foveabox} have been proposed to simplify the detection pipeline. FCOS~\cite{fcos} and CenterNet~\cite{centernet} replace the anchor boxes with anchor points and directly use the points to regress the target objects. CornerNet~\cite{cornernet} first predicts object keypoints and then groups them to bounding boxes using associate embeddings. 

Most CNN-based detectors adopt the one-to-many (o2m) label assignment scheme in the training process. Early detectors such as Faster RCNN~\cite{fasterrcnn}, SSD~\cite{ssd} and RetinaNet~\cite{focalloss} use IoU as a metric to define positive and negative anchors.  FCOS restricts the positive anchor points to be within certain scales and ranges of the object. Recent  methods~\cite{paa,freeanchor,varifocalnet,autoassign,tood,dual} often consider the quality and distribution of the network predictions for more reliable label assignment of anchors. However, the o2m label assignment requires a post-processing step, namely non-maximum suppression (NMS), to remove duplicated predictions. NMS introduces a parameter to compromise precision and recall for all instances, which is however sub-optimal, especially for crowded scenes. In this paper, we aim to remove this hand-crafted NMS step in CNN-based detectors and achieve end-to-end dense detection.

\subsection{Transformer-based Object Detectors}
As a pioneer transformer-based detector, DETR~\cite{detr} utilizes a spare set of learnable object queries as the training candidates to interact with the image feature. It achieves competitive end-to-end detection performance using o2o bipartite matching and the global attention mechanism. However, DETR suffers from slow convergence and inferior performance on small objects. Many following works~\cite{dynamicdetr, acceleratdetr,fastdetr,boxdetr} aim to improve the attention modeling mechanism between the feature map and object queries so that more relevant and precise features can be extracted to boost the performance on small objects. Recent studies~\cite{groupdetr,hybriddetr,dndetr} indicate that it is the limited number of positive samples that slows down the convergence of DETR. Therefore, they introduce several extra decoders to increase the number of positive samples. Nevertheless, these methods are all based on sparse candidates and their computational cost can be unaffordable when performing dense predictions. Different from these methods, we propose a soft label assignment scheme to introduce more positive samples so that end-to-end dense detectors can be more easily trained.
\section{One-to-Few Soft Labeling}
\begin{figure}[tbp]
    \centering
    \includegraphics[width=0.4\textwidth]{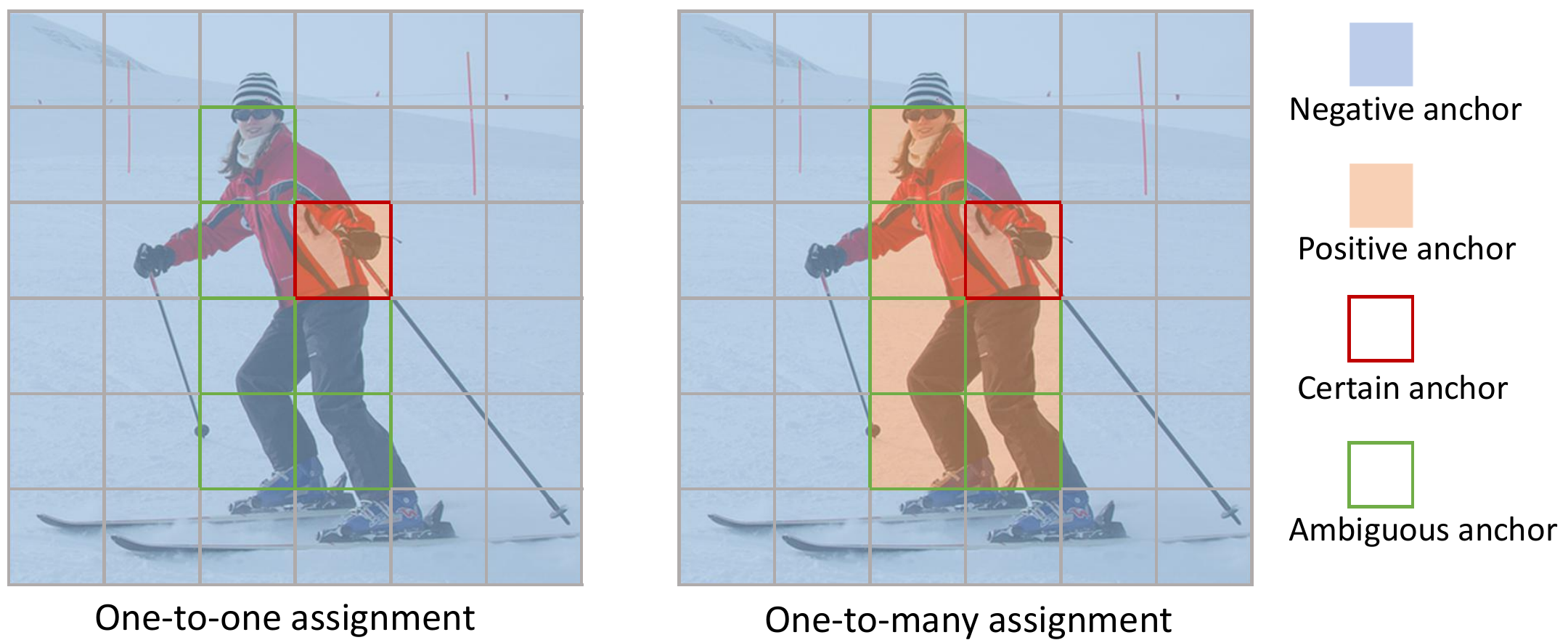}
    \caption{The comparison between o2o label assignment and o2m label assignment. The anchors that have opposite classification labels between o2o and o2m are defined as ambiguous anchors.}
    \label{compare_o2o_o2m}
\end{figure}
\begin{table}[tb]
\centering
\scalebox{1.}{
\setlength{\tabcolsep}{1.5mm}
\begin{tabular}{c|c|c|ccc}
\toprule[1pt]
\multirow{2}{*}{LA} &  \multirow{2}{*}{one-to-one} & \multirow{2}{*}{one-to-two} & \multicolumn{3}{c}{Soft labels} \\
\cline{4-6}
 & & & t=0.1 & t=0.2 & t=0.8\\
 \hline
AP & 36.5 & 25.9 & 37.2 & 37.5 & 25.6 \\
\bottomrule[1pt]
\end{tabular}
}
\caption{Investigation of soft labels for end-to-end detection on COCO. Assigning suitable positive degrees to ambiguous anchors can improve the end-to-end performance. }
\label{table1}
\end{table}
\subsection{Ambiguous Anchors}
Label assignment in dense detection aims to assign each anchor a classification label to supervise the network training. Fig.~\ref{compare_o2o_o2m} illustrates the o2o and o2m label assignments for a ‘person’ instance. One can see that the o2o labeling scheme selects only one anchor as the positive sample, while o2m assigns multiple positive anchors. In both o2o and o2m, the remaining anchors other than positive ones are all defined as negative samples. 

We argue that some anchors actually lie between positive and negative ones, and they should not be simply assigned a single positive or negative label. As shown in Fig.~\ref{compare_o2o_o2m}, we name the anchor (highlighted with red borders) that is positive in both o2o and o2m as the “certain anchor” since there is generally no ambiguity on it. In contrast, we name the anchors (highlighted with green borders) which are positive in o2m but negative in o2o as “ambiguous anchors” as they have conflict labels in o2o and o2m schemes.  

Now we have divided the anchors into three groups: one certain positive anchor, a few ambiguous anchors, and the remaining multiple negative anchors. The ambiguous anchors are labeled as negative in o2o in order to avoid duplicated predictions, whereas they can help learning robust feature representations in o2m. One interesting question is can we find a way to integrate the merits of o2o and o2m schemes so as to improve the performance of end-to-end dense detection? We advocate that the key to solve this problem is how to properly introduce more positive supervision signals into o2o. To find out the solution to this question, let’s test two options first.

The first option is to change o2o to one-to-two by adding one more positive sample for each instance.  The second option is to assign a soft label $t$ to each ambiguous anchor, where $0\leq t \leq 1$ is its positive degree and hence $1-t$ is its negative degree. We define the classification losses of positive and negative anchors as $-log(p)$ and $-log(1-p)$, respectively, where $p$ is the predicted classification score. Then the classification loss of the second option will be $-t\times log(p)-(1-t) \times log(1-p)$. The detection results on the COCO dataset are shown in Table~\ref{table1}, from which we can see that the one-to-two label assignment scheme significantly decreases the performance, even if only one more positive sample is added. In contrast, assigning suitable soft labels to ambiguous anchors can effectively improve the end-to-end performance. (The details of soft label assignment will be discussed in the following sections.)

The above results imply that enabling an ambiguous anchor to be positive and negative simultaneously could be a feasible solution for effective end-to-end dense detection. We therefore propose a one-to-few (o2f) label assignment strategy which selects one certain anchor to be fully positive, a few ambiguous anchors to be both positive and negative, and the remaining anchors to be negative samples. The positive and negative degrees of the ambiguous anchors are dynamically adjusted during the training process so that the network can keep strong feature representation ability and achieve end-to-end detection capability at the same time. 

\begin{figure*}[tbp]
    \centering
    \includegraphics[width=0.9\textwidth]{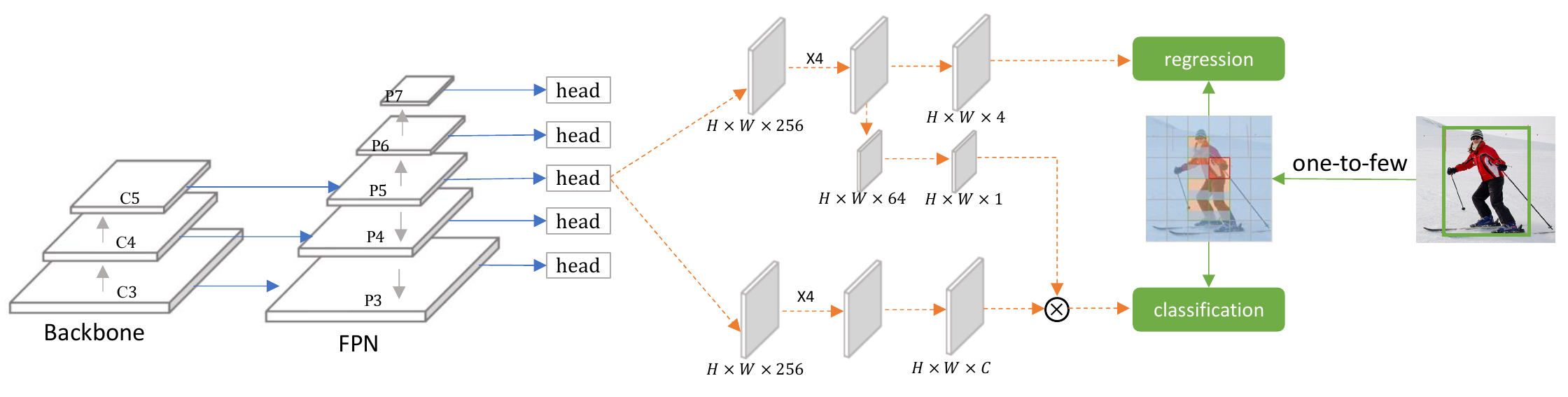}
    \caption{The overall structure of our method. Each FPN layer has a detection head that predicts three outputs: the classification score map of $H \times W \times C$, the regresssion offset map of $H \times W \times 4$, and the centerness/objectness map of $H \times W \times 1$. The structure of the detection head is the same as the one in FCOS except that some extra lightweight convolutional layers with 64 output channels are added before the centerness score map. }
    \label{framework}
\end{figure*}

\subsection{Selection of Certain Positive Anchor}
In our method, a certain positive anchor will be selected for each instance. Previous o2o-based detectors all utilize a prediction-aware selection metric, which considers the cost of both classification and regression to select a unique positive sample. We follow this principle, and incorporate both classification score and IoU into the selection metric for the certain anchor, which is defined as:
\begin{equation}
\begin{aligned}
S_{i,j} = \mathbbm{1}\left[ i \in \Omega _{j} \right] \times p_{i,c_j}^{1-\alpha} \times IoU(b_i,b_j)^\alpha,
\end{aligned}
\label{eq1}
\end{equation}
where  $S_{i,j}$ represents the matching score between anchor $i$ and instance $j$,  $c_j$ is the category label of instance $j$, $p_{i,c_j}$ is the predicted classification score of anchor $i$ belonging to category $c_j$, $b_i$ is the predicted bounding box coordinates of anchor $i$, $b_j$ denotes the coordinates of instance $j$, and $\alpha$ controls the importance degree of classification and regression. $\mathbbm{1}\left[ i \in \Omega _{j} \right]$ is a spatial indicator that outputs 1 when the center point of anchor $i$ is within the central region $\Omega _{j}$  of instance $j$; otherwise it outputs 0. This spatial prior has been commonly utilized in both o2o and o2m methods based on the observation that anchors in the central region of an instance are more likely to be positive ones.

The anchors can be sorted in a descending order according to the metric  $S_{i,j}$. Previous works~\cite{detr,deformdetr} often formulate the positive anchor selection as a bipartite matching problem and solve it by using the Hungarian algorithm~\cite{hungarian}. For simplicity, in this work we directly select the top scored one as the certain positive anchor for each instance. 

\subsection{Label Assignment for Ambiguous Anchors}
Apart from the certain positive anchor, we select the top-$K$ anchors based on the score $S_{i,j}$ as ambiguous anchors since they have similar semantic contexts to the certain positive anchor. To reduce the possibility of duplicated predictions, we assign dynamic soft labels to these ambiguous anchors. Suppose that we train the network for $N$ epochs, the classification loss of each ambiguous anchor $i$ during the $j^{th}$ epoch is defined as:
\begin{equation}
\begin{aligned}
l_i^j = -t_i^j \times log(p_i) - (1-t_i^j) \times log(1-p_i),
\end{aligned}
\label{eq2}
\end{equation}
where $p_i$ is the predicted classification score of anchor $i$, $t_i^j$  and ($1-t_i^j$ ) are the positive and negative degrees (\ie, loss weights) of this anchor at the $j^{th}$ epoch, respectively. $t_i^j$ is dynamically defined as:
\begin{equation}
\begin{aligned}
t_i^j&=\frac{p_i}{\max _k p_k} \times T^j, \\
T^j&=\frac{T^{\min }-T^{\max }}{N-1} \times j+T^{\max },
\end{aligned}
\label{eq3}
\end{equation}
where $T^j$ is a time-dependent variable that is assigned the same value for all samples in the $j^{th}$ epoch, and $T^{max}$ and $T^{min}$ control the positive degree of ambiguous anchors in the first epoch and last epoch, respectively.
We set the loss weights to be positively correlated with the classification scores considering that the anchors with higher prediction scores should contribute more to the positive signals. 
Directly using $p_i$ as the weight will make the training unstable on hard samples because the predicted scores of them are much smaller than that of easy samples. So we use the ratio between $p_i$ and max\{$p$\} to normalize the weights of different samples into the same scale.
Dynamically adjusting $T^j$ is important as it controls the trade-off between `feature learning' and `duplication removal' in different training stages. In the early training stage, we set $T^j$ relatively large to introduce more positive supervision signals for representation learning so that the network can rapidly converge to a robust feature representation space. As the training progresses, we gradually decrease the positive degrees of the ambiguous anchors so that the network can learn to remove duplicated predictions. 

\begin{table*}[tb]
\centering
\scalebox{0.95}{

\begin{tabular}{c|ccccccc}
\toprule[1pt]
$K$ & 13 & 11 & 9 & 7 & 5 & 4 & 3 \\
 \hline
AP & 38.4/38.7 & 38.5/38.8 & 38.6/39.0 & 39.0/39.4 & 38.9/39.3 & 38.6/38.9 & 38.4/38.9 \\
AP50 & 56.0/58.0 & 56.0/58.0 & 56.2/58.0 & 56.7/58.7 & 56.8/58.7 & 56.6/58.5 & 56.3/58.3 \\
AP75 & 42.0/42.0 & 42.0/42.0 & 42.1/42.0 & 42.5/42.5 & 42.4/42.4 & 42.5/42.2 & 42.1/41.8 \\
\bottomrule[1pt]
\end{tabular}
}
\caption{The object detection results on COCO \texttt{val} set by varying the number of ambiguous anchors. The values before and after `/’ denote the results without and with NMS in inference. We set $K$ to 7 in the rest experiments for its best performance.}
\label{table2}
\end{table*}
\subsection{Network Structure}
We instantiate the proposed o2f label assignment strategy to FCOS, which is a typical fully convolutional dense detector. The network structure is shown in Fig.~\ref{framework}. A detection head that consists of two parallel convolutional branches is attached to the output of each Feature Pyramid Network (FPN) layer. One branch predicts a score map of size $H \times W \times C$, where $C$ is the number of categories in the dataset, $H$ and $W$ are the height and width of the feature map, respectively. Another branch predicts a location offsets map of size $H \times W \times 4$ and a centerness map of size $H \times W \times 1$. Following previous works~\cite{poto,autoassign, li2019dynamic}, we multiply the centerness map with the classification score map as the final classification-IoU joint score.

For each instance, we select one certain positive anchor and $K$ ambiguous anchors. The remaining anchors are set as negative samples. The training objective of the classification branch for each instance is formulated as:
\begin{equation}
\begin{aligned}
L_{c l s}=B C E\left(p_c, 1\right)+\sum_{i \in \mathcal{A}} B C E\left(p_i, t_i\right)+\sum_{i \in \mathcal{B}} F L\left(p_i, 0\right),
\end{aligned}
\label{eq4}
\end{equation}
where $p_c$ is the classification score of the single certain anchor, $\mathcal{A}$ and $\mathcal{B}$ represent the set of ambiguous anchors and negative anchors, respectively. BCE indicates the Binary Cross Entropy loss and FL indicates the Focal Loss~\cite{focalloss}. The regression loss is defined as:
\begin{equation}
\begin{aligned}
\begin{matrix}
L_{reg}=\sum_{i \notin \mathcal{B}} GIoU\left(b_i, b_{gt}\right),
\end{matrix}
\end{aligned}
\label{eq5}
\end{equation}
where  GIoU loss is a location loss based on General Intersection over Union~\cite{giou}, $b_i$ is the predicted location of anchor $i$ and $b_{gt}$ is the location of GT object corresponding to anchor $i$. Note that we apply the regression loss on both the positive anchor and the ambiguous anchors.

\section{Experiments}
\subsection{Datasets and Implementation Details}
\textbf{Datasets.} To verify the effectiveness of our o2f method, we conduct experiments on the COCO~\cite{coco} and CrowdHuman~\cite{crowdhuman} datasets. COCO is a challenging benchmark which consists of 118k training images and 5k validation images from 80 classes. We use the standard COCO metric, \ie, AP, by averaging over 10 IoU thresholds ranging from 0.5 to 0.95. CrowdHuman is a widely used dataset for human detection in crowded scenes. It has 15k images for training and 4k images for validation. As suggested by the official paper~\cite{crowdhuman}, mMR, which is the average log miss rate over false positives per-image, is taken as the main metric. In addition, we report the AP and recall results of competing algorithms for reference. 

\textbf{Implementation.} We implement our o2f method by using the MMDetection toolbox~\cite{mmdetection}. All the backbones are initialized by the weights pre-trained on ImageNet~\cite{imagenet} with frozen batch normalizations. The ablation studies are based on ResNet-50~\cite{resnet} backbone with FPN~\cite{fpn}. The input images are resized so that the shorter side is 800 pixels in ‘1$\times$’ schedule. Multi-scale training is applied in ‘3$\times$’ schedule. We train all the models on 8 GPUs with a mini-batch size of 16. When using CNNs as the backbone, we utilize SGDM~\cite{sgd} optimizer with an initial learning rate of 0.01, a momentum of 0.9, and a weight decay of $10^{-4}$. When using Transformers~\cite{swin} as the backbone, we use AdamW~\cite{adamw} with an initial learning rate of 0.0001 and a weight decay of 0.05. In the ‘1$\times$’ schedule, we train the model for 12 epochs and decay the learning rate by a factor of 10 in the $8^{th}$ and $11^{th}$ epochs, respectively. In the ‘3$\times$’ schedule, we train the model for 36 epochs and decay the learning rate in the $24^{th}$ and $33^{th}$ epochs, respectively.

\begin{figure*}[tbp]
    \centering
    \includegraphics[width=0.9\textwidth]{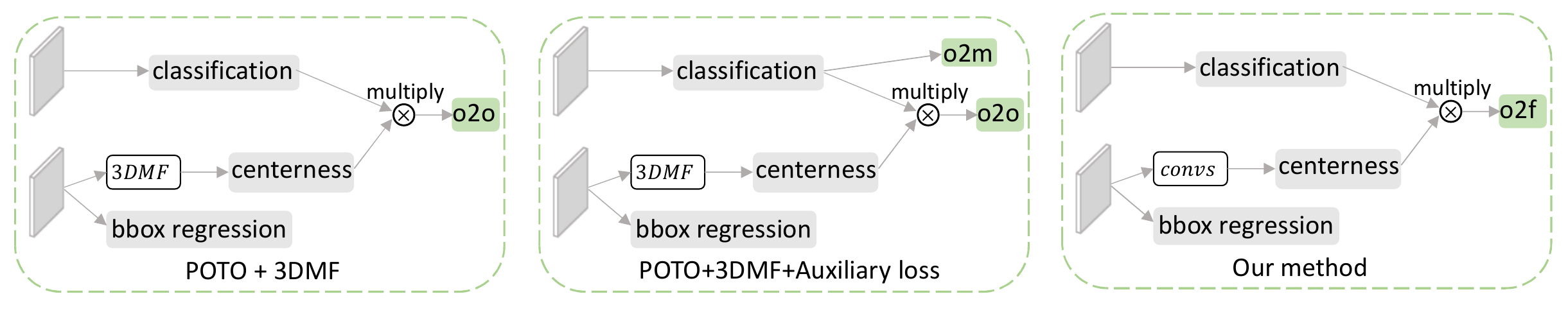}
    \caption{Comparisons of detection head between DeFCN (POTO)~\cite{poto} and our o2f method. In o2f, the classification score and the centerness score are multiplied as the final cls-IoU joint score. DeFCN applies o2o on the joint score and applies o2m on the classification score, while our method applies the dynamic o2f assignment on the joint score. }
    \label{compare_poto_ours}
\end{figure*}
\begin{table}[tb]
\centering
\scalebox{0.95}{

\begin{tabular}{c|c|cccc}
\toprule[1pt]
\multicolumn{2}{c|}{} & \multicolumn{4}{c}{$T^{max}$} \\
\cline{3-6}
 \multicolumn{2}{c|}{} &  0.7 & 0.6 & 0.5 & 0.4\\
 \hline
\multirow{4}{*}{$T^{min}$} & 0.4 & 38.3/39.5 & 38.0/39.5 & 38.0/39.5 & 37.9/39.3 \\
 & 0.3 & 38.8/39.4 & 38.7/39.4 & 38.5/39.4 & 38.6/39.5 \\
 & 0.2 & 38.9/39.3 & 39.0/39.3 & 38.7/39.0 & 38.4/38.8 \\
 & 0.1 & 38.8/39.0 & 38.9/39.1 & 38.6/38.8 & 38.3/38.3 \\
\bottomrule[1pt]
\end{tabular}
}
\caption{The object detection results on COCO \texttt{val} set with different configurations of $T^{max}$ and $T^{min}$. The values before and after ‘/’ denote the results without and with NMS in inference. As shown in the table, a small $T^{min}$ is helpful to narrow down the performance gap between end-to-end detection (without NMS) and non end-to-end detection (with NMS). A large $T^{max}$ can improve the overall performance without NMS. We set $T^{max}$ and $T^{min}$ to 0.6 and 0.2, respectively, in the rest experiments.}
\label{table3}
\end{table}
\begin{table}[tb]
\centering
\scalebox{0.95}{

\begin{tabular}{c|ccc}
\toprule[1pt]
& $\alpha$ & AP & AR \\
 \hline
\multirow{4}{*}{Add} & 0 & 35.3 & 56.7 \\
& 0.6 & 37.9 & 60.5 \\
& 0.8 & 36.7 & 58.5 \\
& 1 & 10.6 & 49.6 \\
\hline
\multirow{4}{*}{Multiply} & 0 & 35.5 & 56.6 \\
& 0.6 & 38.1 & 59.7 \\
& 0.8 & 39.0 & 61.2 \\
& 1 & 10.4 & 49.4 \\
\bottomrule[1pt]
\end{tabular}
}
\caption{The object detection results by using different matching functions in selecting the positive certain anchor and ambiguous anchors. ‘$\alpha=0$' considers the classification score alone, while ‘$\alpha=1$' considers the regression IoU alone. Based on the results, we choose the `Multiply' matching function and set ‘$\alpha=0.8$' in the rest experiments.}
\label{table4}
\vspace{-4mm}
\end{table}
\begin{table*}[tb]
\centering
\scalebox{0.95}{
\setlength{\tabcolsep}{0.8mm}
\begin{tabular}{c|c|ccccc|ccc}
\toprule[1pt]
Method & LA & AP(1$\times$) & AR(1$\times$) & AP(3$\times$) & AR(3$\times$) & NMS & forward(ms) & NMS(ms) & FPS \\
 \hline
FCOS~\cite{fcos} & o2m & 38.6 & 57.2 & 41.4 & 59.1 & \Checkmark & 27.7 & 0.7 & 35.2 \\
FCOS~\cite{fcos} & o2m & 17.7 & 52.9 & 19.1 & 54.2 & \XSolidBrush & 27.7 & 0 & 36.1 \\
POTO~\cite{poto} & o2o & 36.5 & 58.9 & 40.2 & 61.1 & \XSolidBrush & 27.7 & 0 & 36.1 \\
POTO+3DMF~\cite{poto} & o2o & 37.0 & 58.8 & 40.5 & 60.9 & \XSolidBrush & 30.3 & 0 & 33.0 \\
POTO+3DMF+Aux~\cite{poto} & o2o+o2m(ATSS) & 37.6 & 58.7 & 41.2 & 61.2 & \XSolidBrush & 30.3 & 0 & 33.0 \\
POTO+3DMF+Aux~\cite{poto} & o2o+o2m(Top-k) & 37.6 & 59.0 & 41.1 & 61.3 & \XSolidBrush & 30.3 & 0 & 33.0 \\
POTO+3DMF+Aux~\cite{poto} & o2o+o2m(FCOS) & 36.5 & 58.0 & 40.3 & 60.6 & \XSolidBrush & 30.3 & 0 & 33.0 \\
Ours-1conv & o2f & 38.4 & 60.5 & 41.6 & 63.3 & \XSolidBrush & 28.1 & 0 & 35.6 \\
Ours-2convs & o2f & 38.7 & 60.6 & 42.0 & 63.5 & \XSolidBrush & 28.2 & 0 & 35.5 \\
Ours-3convs & o2f & 39.0 & 61.2 & 42.2 & 63.5 & \XSolidBrush & 28.2 & 0 & 35.5 \\
\bottomrule[1pt]
\end{tabular}
}
\caption{Comparisons with state-of-the-art end-to-end dense detectors on COCO \texttt{val} set. All experiments are conducted with ResNet-50 backbone. `LA' means label assignment. `o2o' means one-to-one. `o2m' means one-to-many. `o2f' means one-to-few. `ATSS', `Top-k' and `FCOS' in the `LA' column are the different o2m label assignment strategies used in the auxiliary loss in POTO. The reported runtime (ms) are all evaluated on a Tesla-V100 GPU under the MMDetection toolbox.}
\label{table5}
\end{table*}
\subsection{Ablation Studies}
\textbf{The number of ambiguous anchors.} In our o2f method, we select $K$ ambiguous anchors that can have both positive and negative losses. We ablate this hyper-parameter $K$ in Tab.~\ref{table2} on COCO \texttt{val} set. One can see that the best AP performance 39.0 (without NMS) is obtained when $K$ equals to 7. We set $K$ to 7 in the rest experiments.

\textbf{$\boldsymbol{T^{max}}$ and $\boldsymbol{T^{min}}$.} These two parameters control the positive loss weights of ambiguous anchors in the first and last epoch, respectively,  during the training process. Tab.~\ref{table3} shows the results on COCO \texttt{val} set by using different combinations of $T^{max}$ and $T^{min}$. We can see that when $T^{max}$ is set to a higher value like 0.6, increasing $T^{min}$ will decrease the end-to-end detection performance without NMS. This implies that a small $T^{min}$ is helpful to narrow down the performance gap between ‘without NMS’ and ‘with NMS’. Meanwhile, when $T^{min}$ is set to a smaller value like 0.1, a small $T^{max}$ can decrease the performance with and without NMS due to the limited number of positive samples. The best AP performance 39.0 is achieved when we set $T^{max}$ and $T^{min}$ to 0.6 and 0.2, respectively, which is the default setting in our experiments.

\textbf{Selection metric.} We further explore the effect of matching functions in selecting the certain positive anchor and the ambiguous anchors. As shown in Tab.~\ref{table4}, the method `Multiply' means the matching function in Eq.~\ref{eq1}, while the method ‘Add’ means that we change Eq.~\ref{eq1} to  $(1-\alpha)\times p_{i,c_j}+\alpha  \times IoU(b_i,b_j)$, which is a frequently used metric in DETRs~\cite{deformdetr,detr}. We see that considering only the classification score (\ie, $\alpha=0$) can decrease the performance without NMS. Considering only the regression IoU (\ie, $\alpha=1$)  can result in duplicated predictions and significantly enlarge the gap with NMS. We can also see that ‘Multiply’ is more suitable for dense detection which achieves 1.1 points gains over the best ‘Add’ result. We choose ‘Multiply’ and set $\alpha=0.8$ in the rest experiments. 

\subsection{Results on COCO}
We mainly compare our o2f method with DeFCN~\cite{poto}, which is the representative and leading o2o method for fully convolutional dense detection. DeFCN employs a Prediction-aware One-to-One (POTO) label assignment scheme for classification and holds the state-of-the-art end-to-end dense detection performance. The framework comparison between DeFCN and our o2f is shown in Fig.~\ref{compare_poto_ours}. DeFCN adds an auxiliary o2m loss for the classification score in addition to the standard o2o loss. It also introduces a 3D Max Filtering (3DMF) module in the centerness branch to increase the local discrimination ability of convolutions. In contrast, we apply a single dynamic o2f label assignment loss in the final loss function and introduce several lightweight convolutional layers in the centerness branch with negligible computation overhead.

\textbf{Main results.} Tab.~\ref{table5} presents the detailed comparisons between POTO and our method in both detection accuracy and speed. The original FCOS uses o2m label assignment during training, achieving 38.6 AP/17.7 AP with/without NMS in inference by using ‘1$\times$’ schedule. Such a huge gap between `with' and without `NMS' indicates that o2m can easily produce duplicated predictions. By replacing o2m with o2o during training, POTO significantly improves the end-to-end performance from 17.7 to 36.5, demonstrating that o2o plays a crucial role in removing duplicated predictions. POTO further increases the performance to 37.6 by adding 3DMF and the auxiliary loss. However, the o2m label assignment in the auxiliary loss has to be carefully tuned. Using the FCOS-style o2m strategy in the auxiliary loss can decrease the performance. What’s more, DeFCN increases the forward time from 27.7 ms to 30.3 ms and runs slower than the original FCOS (33.0 FPS vs 35.2 FPS).

Our o2f method achieves 39.0 AP in `1$\times$' schedule when using three extra convolution layers in the centerness branch, which achieves 1.4 points gains (39.0 AP vs. 37.6 AP) and a faster inference speed over DeFCN (35.5 FPS vs. 33.0 FPS). It is worth mentioning that our method also achieves better accuracy and faster inference speed than FCOS with NMS. 


A recent similar approach to ours is the Hybrid-Epoch scheme in H-DETR~\cite{hybriddetr}, which can also be seen as a combination of o2o and o2m. The key difference lies in that H-DETR uses static loss weights, where the weights of o2o and o2m are \{0,1\} in the early training epochs and \{1,0\} in the later training epochs. In contrast, we use a single soft LA branch and dynamically adjust the loss weights. We implement the Hybrid-Epoch scheme on FCOS and achieve 37.3 and 40.6 AP under the 1$\times$ and 3$\times$ schedules, respectively, which are 1.7 and 1.6 points lower than our o2f. This shows that static loss weights are less effective than dynamic loss weights for end-to-end dense detection. 

\begin{figure}[tbp]
    \centering
    \includegraphics[width=0.4\textwidth]{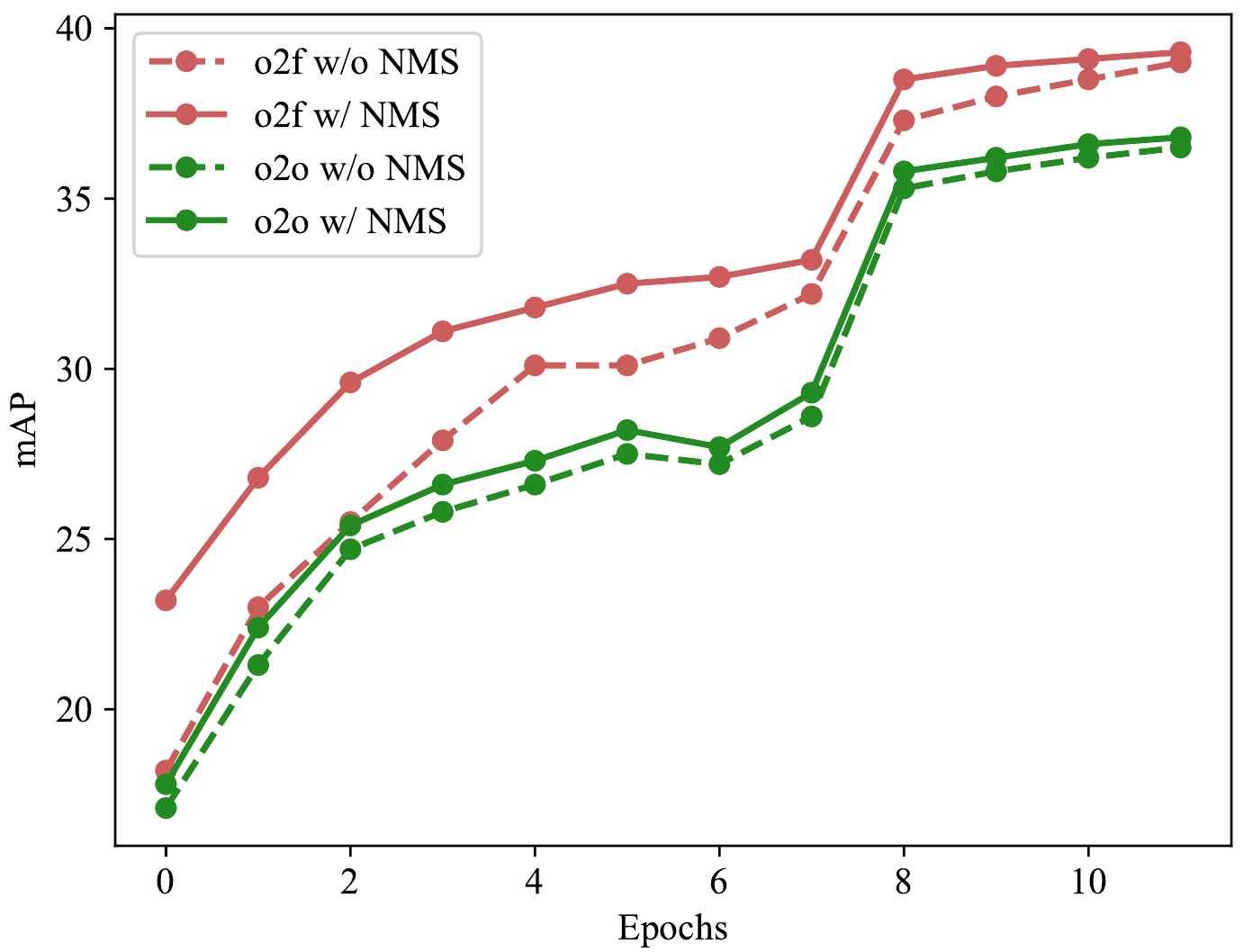}
    \caption{The object detection performance $w.r.t.$ training epochs for o2o and o2f methods. All models are based on ResNet-50 backbone and `1$\times$' schedule. The threshold in NMS is 0.6.}
    \label{ap_curve}
    \vspace{-4mm}
\end{figure}

\textbf{Model performance during training process.}
We show in Fig.~\ref{ap_curve} the performance of our model during the training process. One can see that at the very beginning, the gap between `w/o NMS' and `w/ NMS' is large because we assign relatively large positive weights to the ambiguous anchors. As the training progresses, we gradually decrease the positive loss weights of the ambiguous anchors and thus the gap becomes smaller and smaller. This phenomenon conforms to our motivation that we focus on `feature representation learning' in the early training stage and `duplicated prediction removal' in the later training stage. In contrast, the model trained with the o2o label assignment keeps a small gap during the whole training process, but finally leads to inferior performance due to the limited number of positive training samples.

\textbf{Larger backbones.}
To further demonstrate the robustness and effectiveness of our method, we provide experiments with larger backbones. The detailed results are shown in Tab.~\ref{res101} and Tab.~\ref{swin}. When using ResNet-101 as the backbone, our method performs better than POTO (o2o) by 2.3 points and 1.8 points under the `1$\times$’ and `3$\times$’ schedules, respectively. When using Swin-T  as the backbone, our method achieves 2.1 and 2.1 AP gains over POTO under the the `1$\times$' and `3$\times$' schedules, respectively.

\begin{figure*}[tbp]
    \centering
    \includegraphics[width=0.85\textwidth]{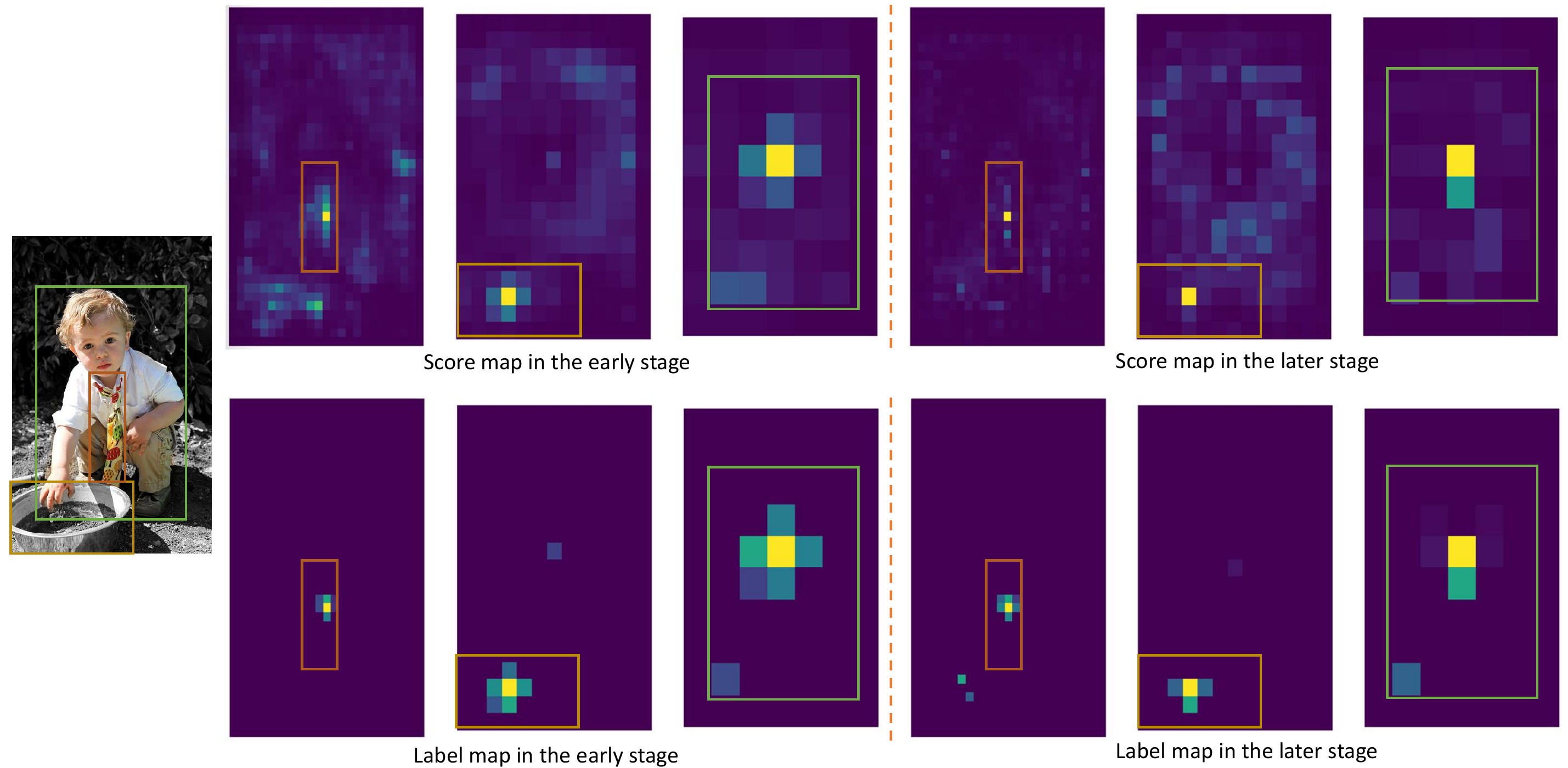}
    \caption{Visualization of the predicted classification scores and assigned soft labels in the early and later training stages. There are three instances of different scales in the input image, \ie, person, tie and pot. The heatmaps from the left to right in each stage correspond to the FPN layers of `P5', `P6' and `P7', respectively.}
    \label{map_vis}
\end{figure*}

\textbf{Visualization.} In Fig.~\ref{map_vis}, we visualize the classification scores and assigned labels in the early and later training stages of our o2f method. From left to right, the three maps in each stage correspond to the instances of `tie', `pot' and `person', respectively. In the early training stage, we assign relatively large positive weights to the ambiguous anchors so that they contribute more to feature representation learning. This causes duplicated predictions as shown in the score map, where several anchors are highly activated for one instance. In the later training stage, we gradually decrease the positive weights of the ambiguous anchors, and the classification score maps become much sparser, which demonstrates that the network has learned to remove the duplicated predictions. The above results indicate that to achieve end-to-end detection, there is no need to enforce sparse predictions at the early training stage, where `feature learning' should be paid more attention.

\begin{table}[tb]
\centering
\scalebox{0.85}{
\setlength{\tabcolsep}{0.8mm}
\begin{tabular}{c|cccccc}
\toprule[1pt]
Method & AP(1$\times$) & AR(1$\times$) & AP(3$\times$) & AR(3$\times$) & NMS \\
 \hline
FCOS~\cite{fcos} & 41.0 & 59.7 & 43.5 & 60.0 & \Checkmark \\
FCOS~\cite{fcos} & 18.6 & 54.9 & 21.2 & 56.1 & \XSolidBrush \\
POTO~\cite{poto} & 38.6 & 60.2 & 42.0 & 62.9 & \XSolidBrush \\
POTO+3DMF+Aux~\cite{poto} & 39.8 & 60.3 & 42.7 & 61.4 & \XSolidBrush \\
Ours & 40.9 & 62.3 & 43.8 & 64.9 & \XSolidBrush \\
\bottomrule[1pt]
\end{tabular}
}
\caption{The object detection results with ResNet-101 backbone.}
\label{res101}
\end{table}
\begin{table}[tb]
\centering
\scalebox{0.85}{
\setlength{\tabcolsep}{0.8mm}
\begin{tabular}{c|ccccc}
\toprule[1pt]
Method & AP(1$\times$) & AR(1$\times$) & AP(3$\times$) & AR(3$\times$) & NMS \\
 \hline
FCOS~\cite{fcos} & 43.6 & 61.1 & 46.8 &  63.7& \Checkmark \\
FCOS~\cite{fcos} & 20.7 & 57.2 & 23.0 & 59.6 & \XSolidBrush \\
POTO~\cite{poto} & 40.8 & 61.6 & 44.0 & 64.7 & \XSolidBrush \\
POTO+3DMF+Aux~\cite{poto} & 41.8 & 61.7 &  44.8 & 63.9 & \XSolidBrush \\
Ours & 42.9 & 63.7 & 46.1 & 66.7  & \XSolidBrush \\
\bottomrule[1pt]
\end{tabular}
}
\caption{The object detection results with Swin-T backbone.}
\vspace{-4mm}
\label{swin}
\end{table}
\begin{table}[tb]
\centering
\scalebox{0.9}{
\setlength{\tabcolsep}{1mm}
\begin{tabular}{c|ccccc}
\toprule[1pt]
Method & AP $\uparrow$ & mAR $\downarrow$ & Recall $\uparrow$ & NMS \\
 \hline
FCOS~\cite{fcos} & 86.1 & 55.2 & 94.3 & \Checkmark \\
POTO~\cite{poto} & 88.7 & 52.0 & 96.6 & \XSolidBrush \\
POTO+3DMF+Aux~\cite{poto} & 89.2 & 49.6 & 96.6 & \XSolidBrush \\
Ours & 90.9 & 45.2 & 97.9 & \XSolidBrush \\
\bottomrule[1pt]
\end{tabular}
}
\caption{Experiments on CrowdHuman.}
\label{crowdhuman}
\vspace{-2mm}
\end{table}
\subsection{Results on CrowdHuman}
To further validate the generalization ability of our method, we conduct experiments on CrowdHuman, which is a widely used dataset for human detection in crowded scenes. NMS-based methods use an IoU threshold to filter out duplicated predictions, which suffers from a dilemma: a high threshold can bring more false positives while a low threshold can suppress true positives. This problem becomes more severe in crowed scenes. Our proposed end-to-end dense detector is free of NMS and hence it can avoid this problem. The results are shown in Tab.~\ref{crowdhuman}. One can see that our o2f method significantly outperforms the NMS-based FCOS by 10 points in mAR, which is the main metric in crowed detection. It also outperforms POTO by 6.8 points in mAR, clearly demonstrating the advantages of our o2f strategy over o2o and o2m strategies. 

\subsection{Results on Instance Segmentation}
Considering that the framework of instance segmentation shares similarities with object detection and it also employs NMS to suppress duplicated prediction masks, we apply the proposed o2f label assignment strategy to the popular single-stage dense instance segmentation method CondInst~\cite{condinst} to validate its effectiveness. The results based on ResNet-50 backbone with `1$\times$' and `3$\times$' learning schedules are shown in Tab.~\ref{condinst}.
The original CondInst trained with o2m strategy has very poor performance without using NMS in inference. Using o2o during training can effectively improve the end-to-end segmentation performance to 32.8 AP and 36.1 AP under the `1$\times$' and `3$\times$' schedules, respectively. Our o2f label assignment strategy can further boost the performance by 3.1 and 1.9 points, respectively. 
Visual results can be found in the \textbf{supplementary file.}
\begin{table}[tb]
\centering
\scalebox{0.9}{
\setlength{\tabcolsep}{0.8mm}
\begin{tabular}{c|cccccc}
\toprule[1pt]
Method & AP(1$\times$) & AR(1$\times$) & AP(3$\times$) & AR(3$\times$) & NMS \\
 \hline
CondInst~\cite{condinst}  & 35.4  &  51.7 & 38.1 & 53.6 & \Checkmark \\
CondInst~\cite{condinst} & 9.5 & 45.1 & 10.7 & 47.5 & \XSolidBrush \\
POTO~\cite{poto} &  32.8 &  52.2 & 36.1 & 54.6 & \XSolidBrush \\
Ours & 35.9 & 53.9 & 38.0 &  61.3& \XSolidBrush \\
\bottomrule[1pt]
\end{tabular}
}
\caption{The instance segmentation results on COCO val set. All models are based on ResNet-50.}
\vspace{-3mm}
\label{condinst}
\end{table}

\section{Conclusion}
We presented a novel one-to-few (o2f) label assignment strategy to achieve fully convolutional end-to-end dense detection without NMS postprocessing. By assigning soft labels to ambiguous anchors and dynamically adjusting their positive and negative degrees in the training process, we enabled the detectors strong feature representation and duplicated prediction removal abilities at the same time. The proposed o2f method exhibited superior performance to state-of-the-art end-to-end dense detectors on the COCO benchmark. Its advantages became more obvious on crowed scenes, as we demonstrated on the CrowdHuman dataset. In addition, our proposed o2f method can also be applied to detection related tasks such as instance segmentation. 
\section{Supplementary Materials}
In this supplementary file, we present more visualization results on COCO detection, CrowdHuman detection and COCO instance segmentation.

\begin{figure}[htb]
	\centering

	\includegraphics[width = .3\linewidth]{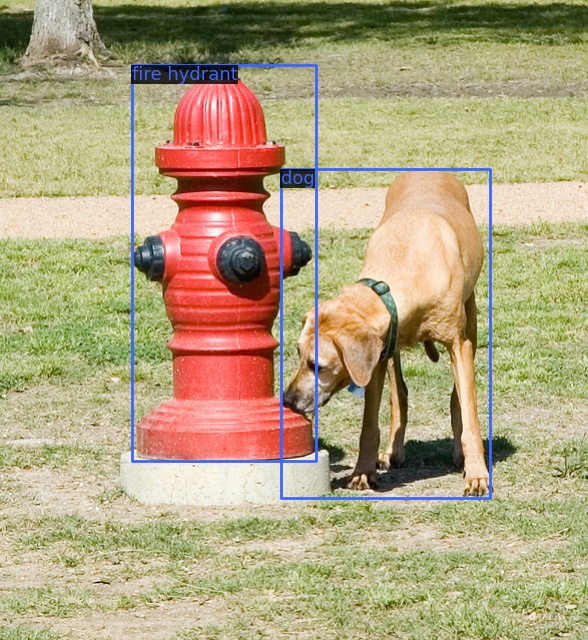}
	\includegraphics[width = .3\linewidth]{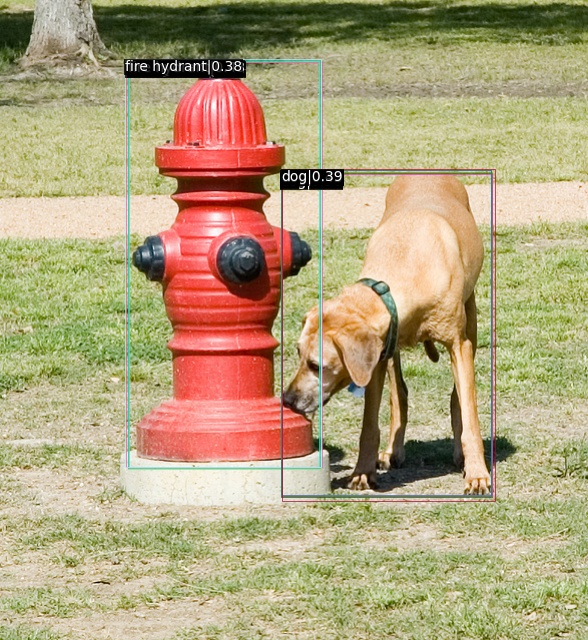}
	\includegraphics[width = .3\linewidth]{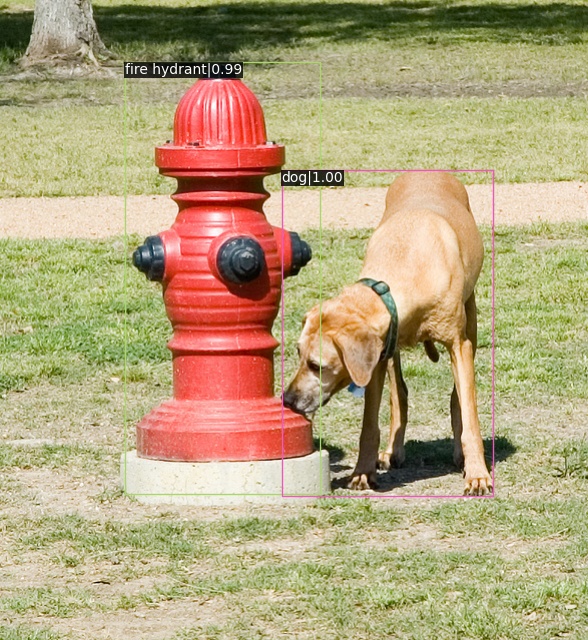}
	
	\includegraphics[width = .3\linewidth]{supp_figures/coco_det/gt/000000292155.jpg}
	\includegraphics[width = .3\linewidth]{supp_figures/coco_det/fcos_wonms/000000292155.jpg}
	\includegraphics[width = .3\linewidth]{supp_figures/coco_det/o2f/000000292155.jpg}
	
	\includegraphics[width = .3\linewidth]{supp_figures/coco_det/gt/000000221693.jpg}
	\includegraphics[width = .3\linewidth]{supp_figures/coco_det/fcos_wonms/000000221693.jpg}
	\includegraphics[width = .3\linewidth]{supp_figures/coco_det/o2f/000000221693.jpg}
	
	\includegraphics[width = .3\linewidth]{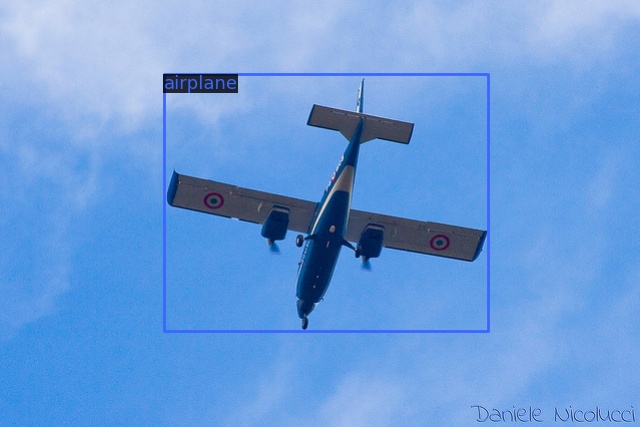}
	\includegraphics[width = .3\linewidth]{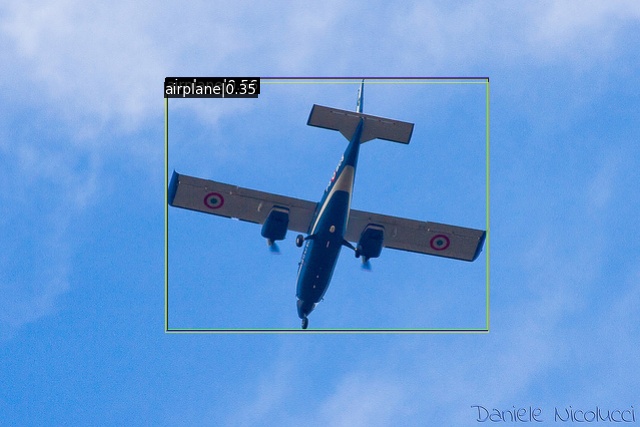}
	\includegraphics[width = .3\linewidth]{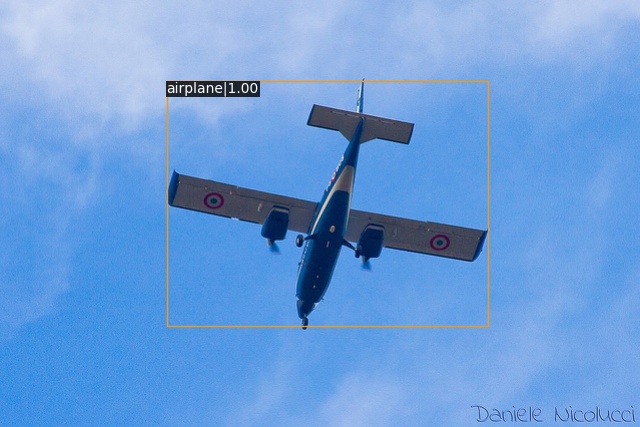}
	
	\subcaptionbox{Ground-truth}{\includegraphics[width = .3\linewidth]{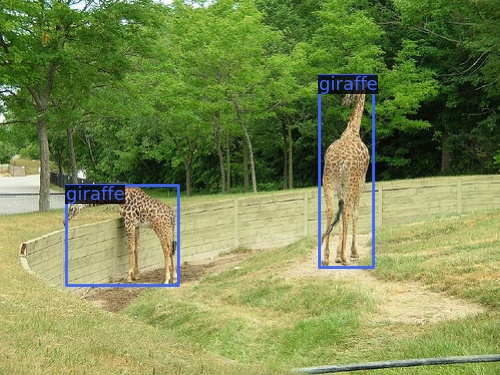}}
	\subcaptionbox{FCOS w/o NMS}{\includegraphics[width = .3\linewidth]{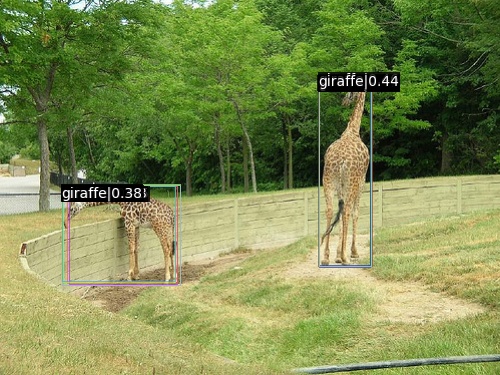}}
	\subcaptionbox{Ours}{\includegraphics[width = .3\linewidth]{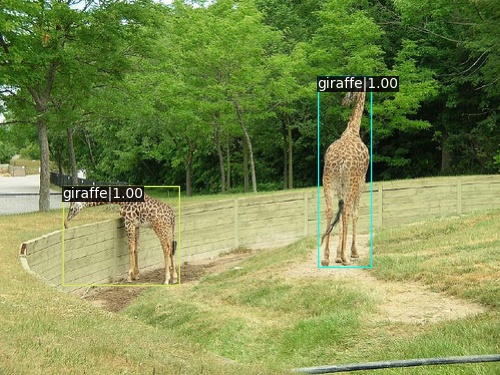}}
	\caption{The prediction visualizations of different detectors on COCO val set.}
	\label{fig7}
\end{figure}

Fig.~\ref{fig7} shows the detection results of FCOS and our method on COCO val set. The official FCOS is trained with one-to-many label assignment. We can clearly see that such a strategy can cause duplicated predictions when NMS is not used in inference. In contrast, our method can directly predict one unique bounding box for one object without using NMS.

\begin{figure}[htb]
	\centering
	\includegraphics[width = .3\linewidth, height=0.2\linewidth]{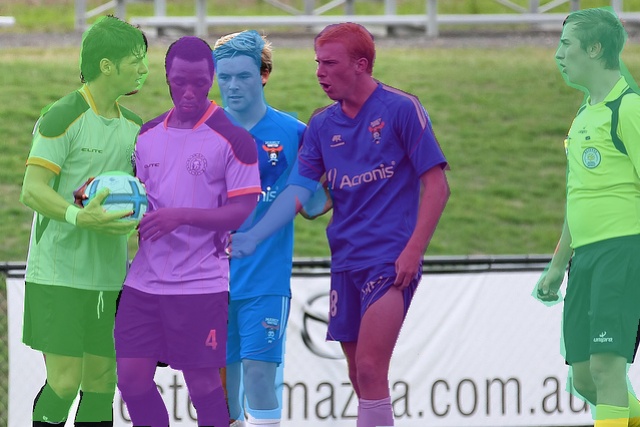}
	\includegraphics[width = .3\linewidth,height=0.2\linewidth]{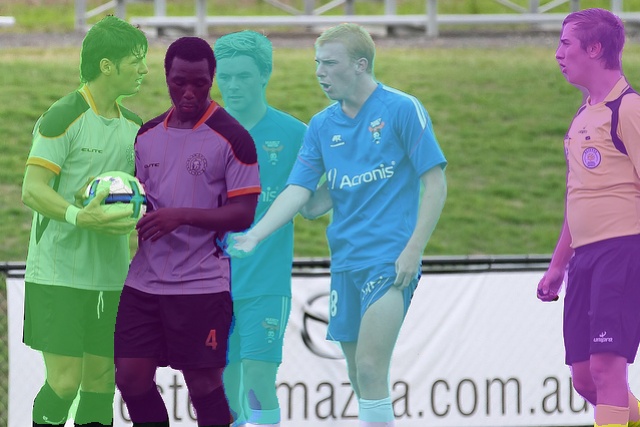}
	\includegraphics[width = .3\linewidth,height=0.2\linewidth]{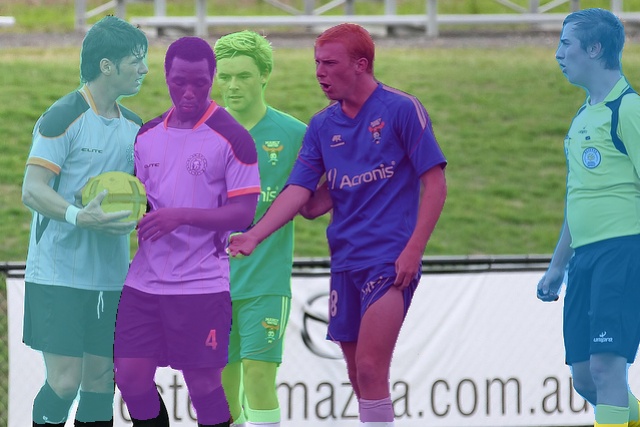}
	
	\includegraphics[width = .3\linewidth,height=0.2\linewidth]{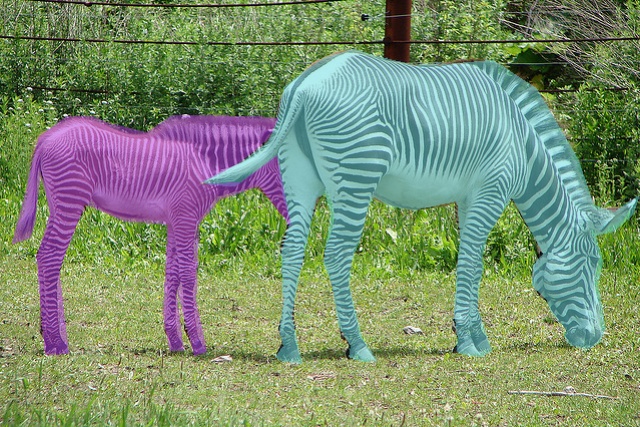}
	\includegraphics[width = .3\linewidth,height=0.2\linewidth]{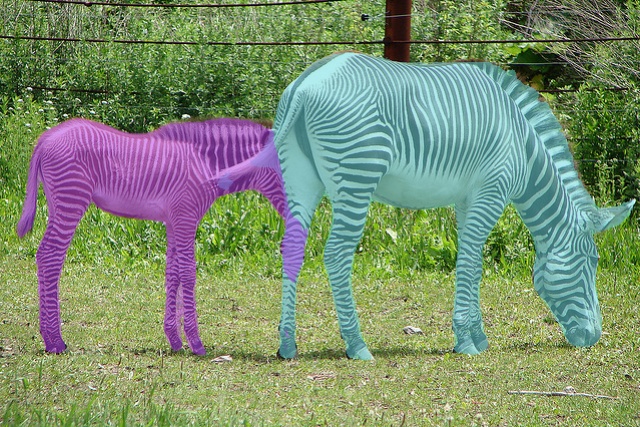}
	\includegraphics[width = .3\linewidth,height=0.2\linewidth]{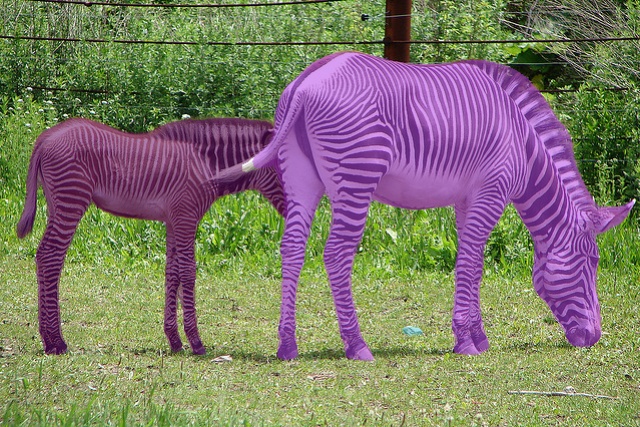}
	
	\includegraphics[width = .3\linewidth,height=0.2\linewidth]{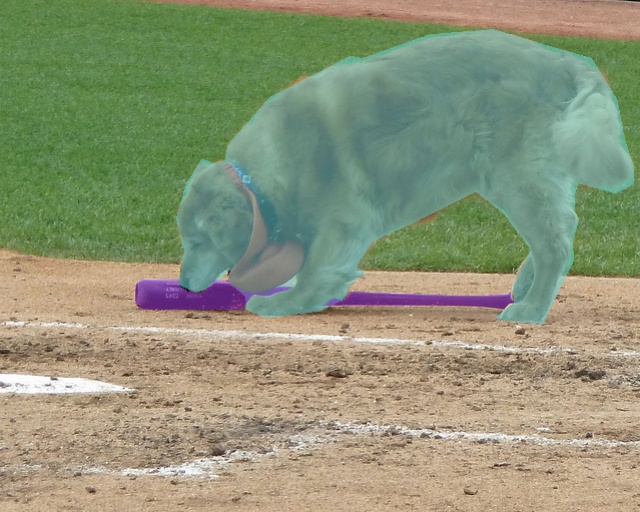}
	\includegraphics[width = .3\linewidth,height=0.2\linewidth]{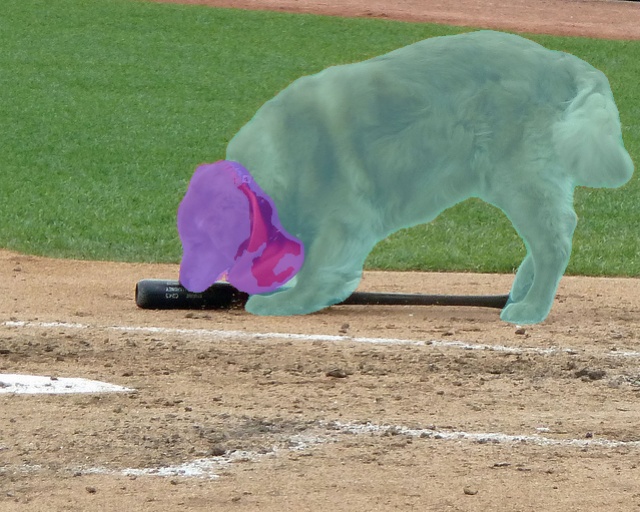}
	\includegraphics[width = .3\linewidth,height=0.2\linewidth]{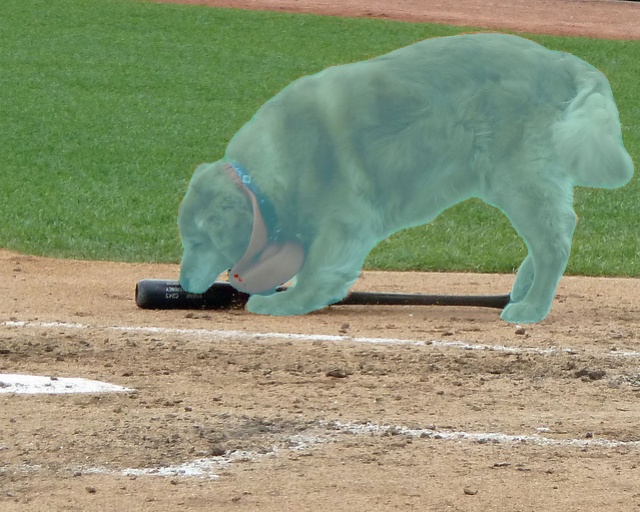}
	
	\includegraphics[width = .3\linewidth,height=0.2\linewidth]{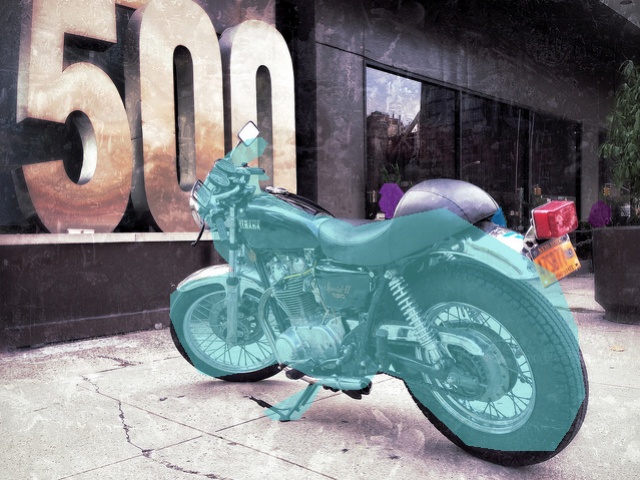}
	\includegraphics[width = .3\linewidth,height=0.2\linewidth]{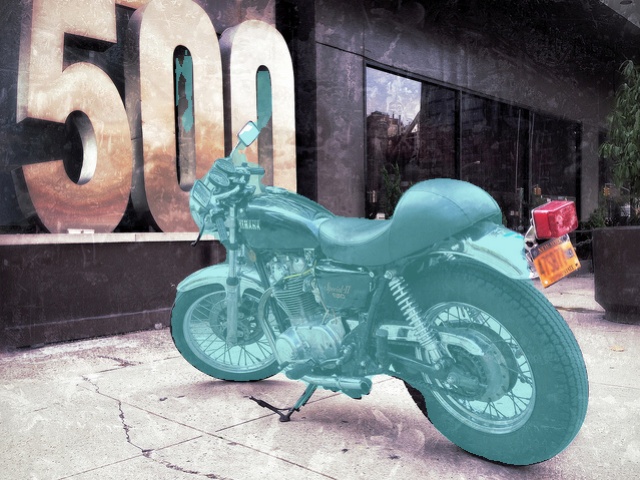}
	\includegraphics[width = .3\linewidth,height=0.2\linewidth]{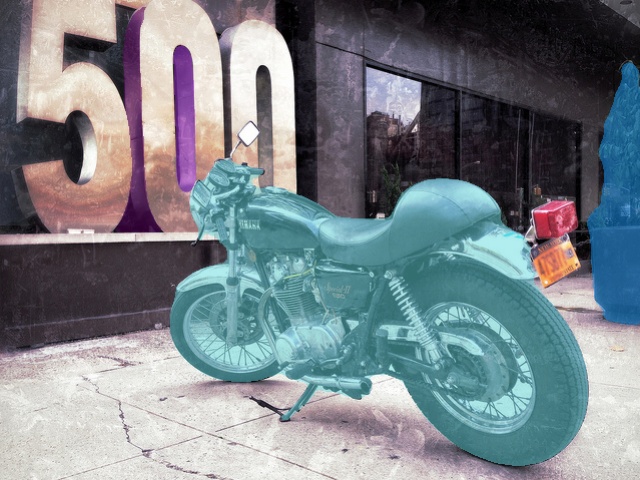}
	
	\subcaptionbox{Ground-truth}{\includegraphics[width = .3\linewidth,height=0.2\linewidth]{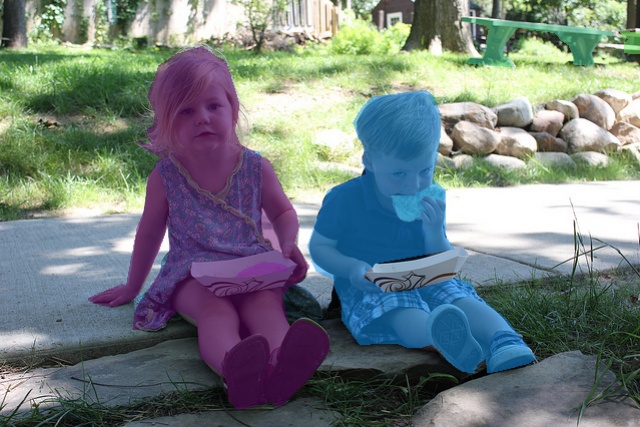}}
	\subcaptionbox{POTO}{\includegraphics[width = .3\linewidth,height=0.2\linewidth]{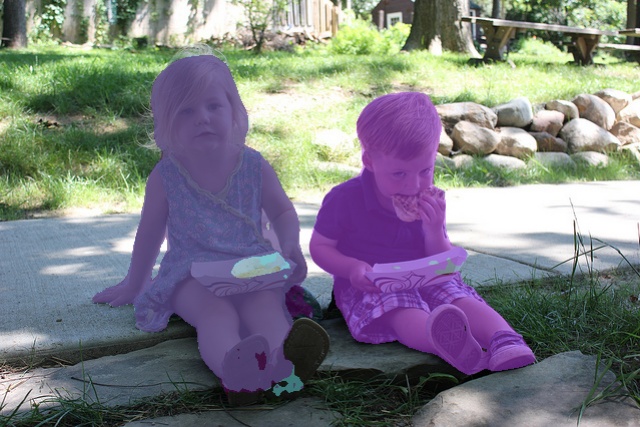}}
	\subcaptionbox{Ours}{\includegraphics[width = .3\linewidth,height=0.2\linewidth]{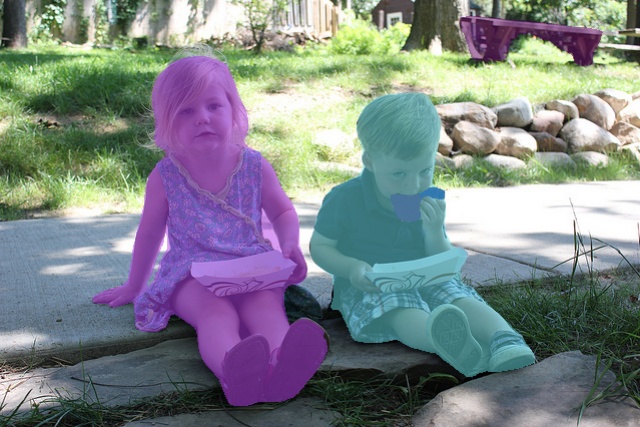}}
	
	\caption{The prediction visualizations of different instance segmentation methods on COCO val set.}
	\label{fig8}
	\vspace{2mm}
\end{figure}

Fig.~\ref{fig8} shows the instance segmentation results of POTO and our method on COCO val set. POTO adopts one-to-one label assignment strategy during training and it can avoid NMS in inference but it predicts unsatisfied masks, especially in the overlapped regions as shown in Fig.~\ref{fig8}. Thanks to the dynamic soft label assignment strategy, our method is able to predict one \textbf{unique} and \textbf{precise} mask for one object.

\begin{figure}[htb]
	\centering

	\includegraphics[width = .3\linewidth]{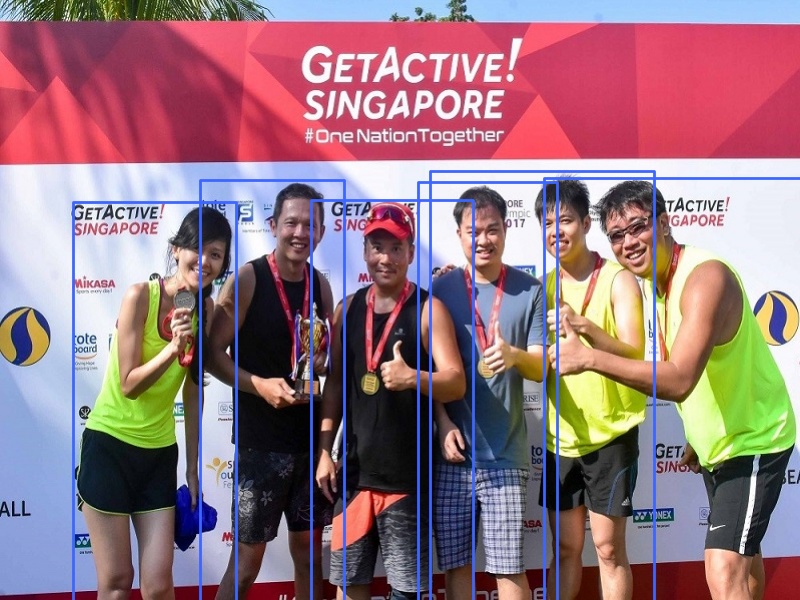}
	\includegraphics[width = .3\linewidth]{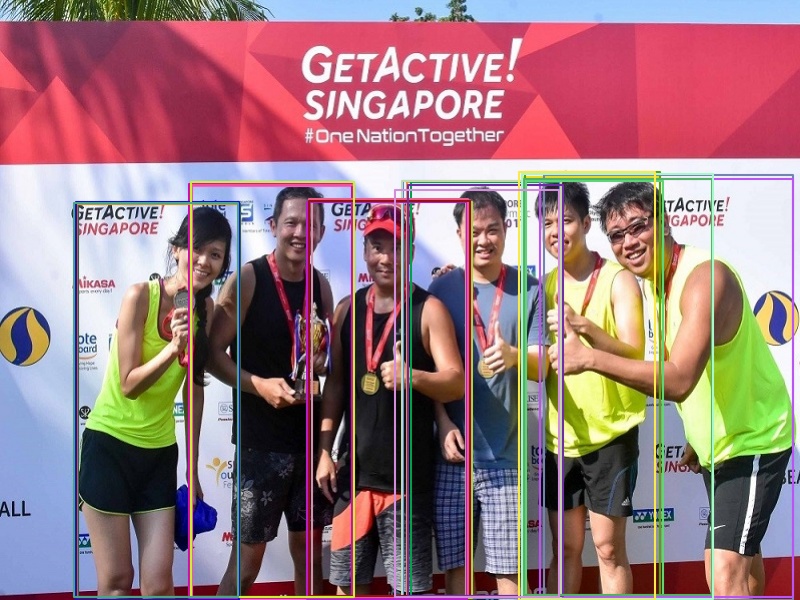}
	\includegraphics[width = .3\linewidth]{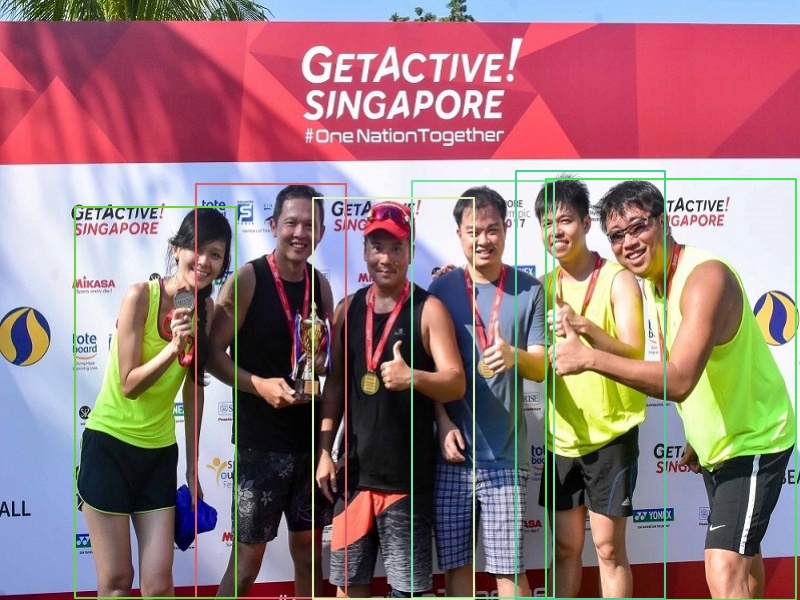}
	
	\includegraphics[width = .3\linewidth]{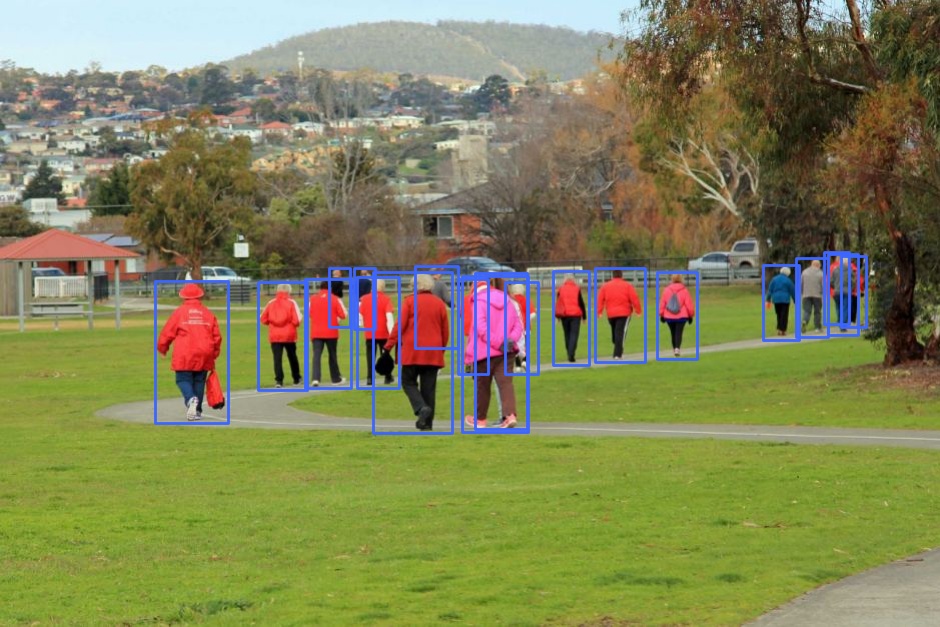}
	\includegraphics[width = .3\linewidth]{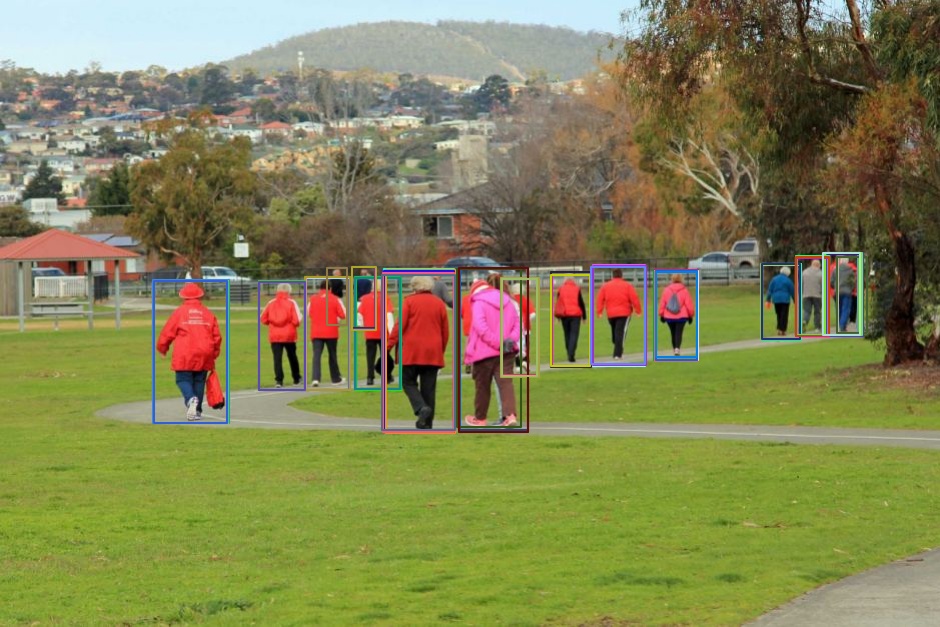}
	\includegraphics[width = .3\linewidth]{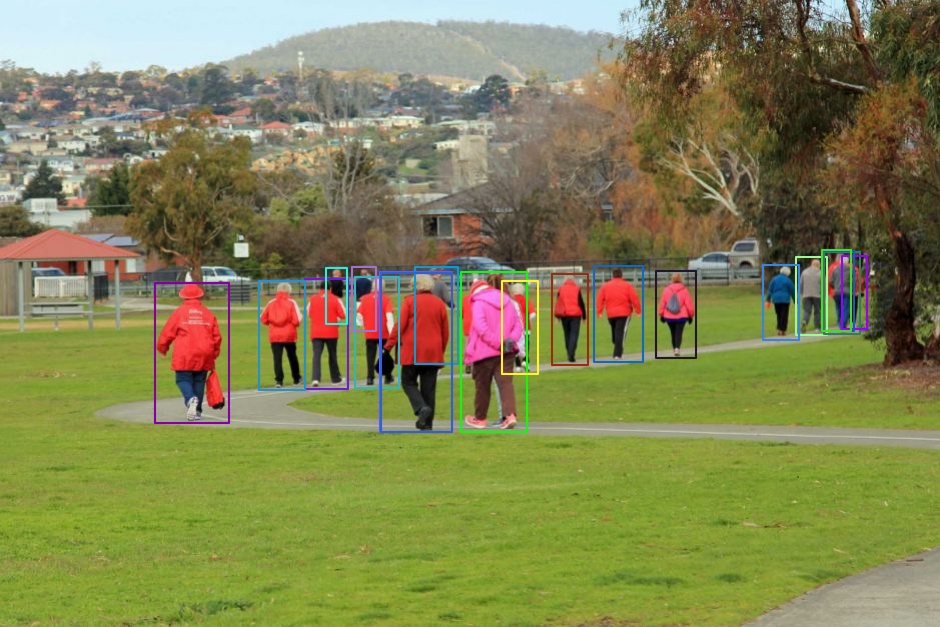}
	
	\includegraphics[width = .3\linewidth]{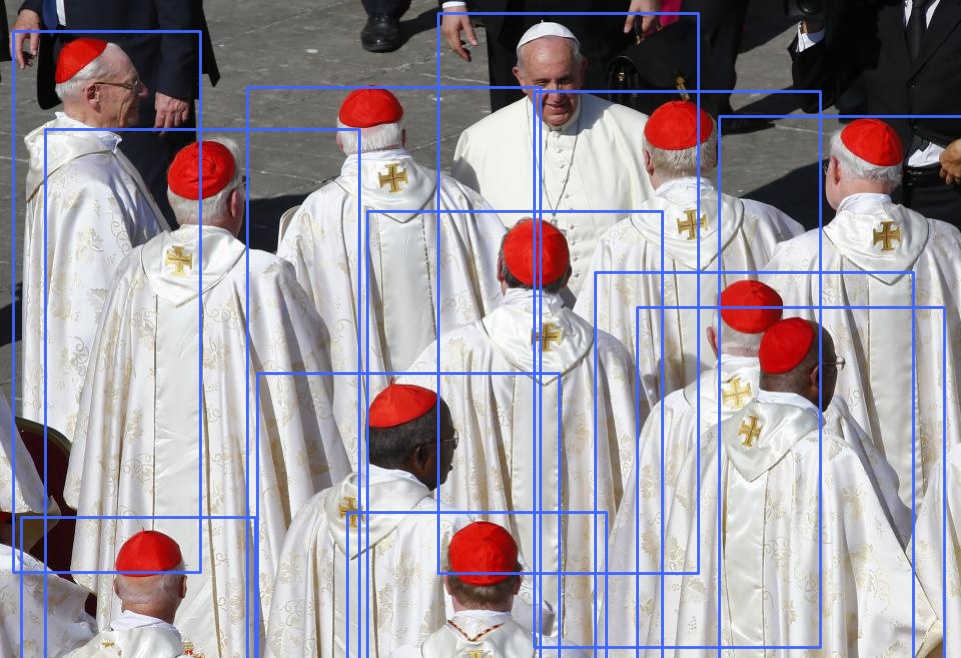}
	\includegraphics[width = .3\linewidth]{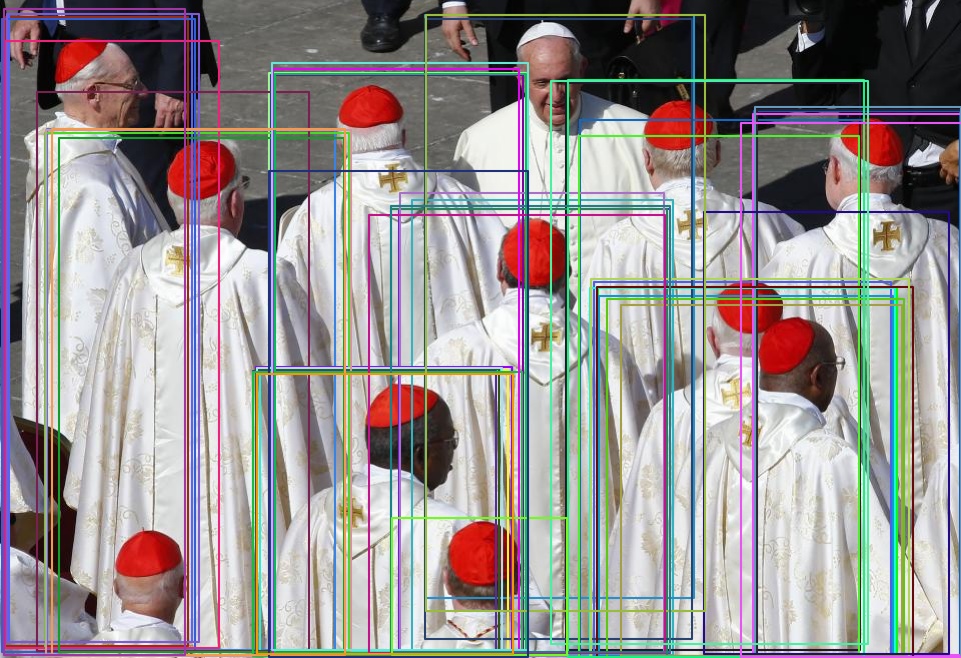}
	\includegraphics[width = .3\linewidth]{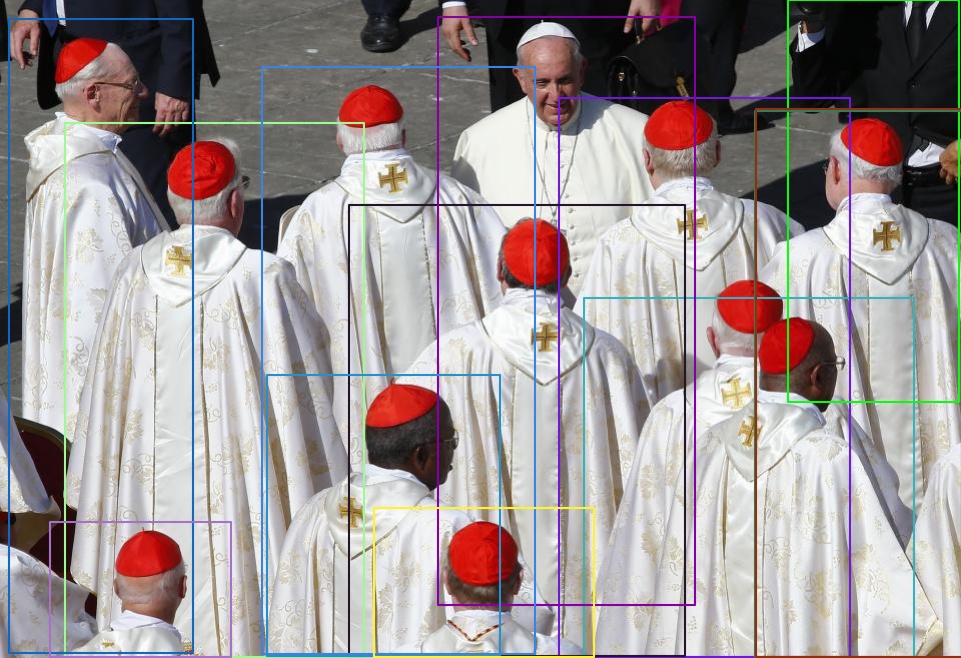}
	
	\includegraphics[width = .3\linewidth]{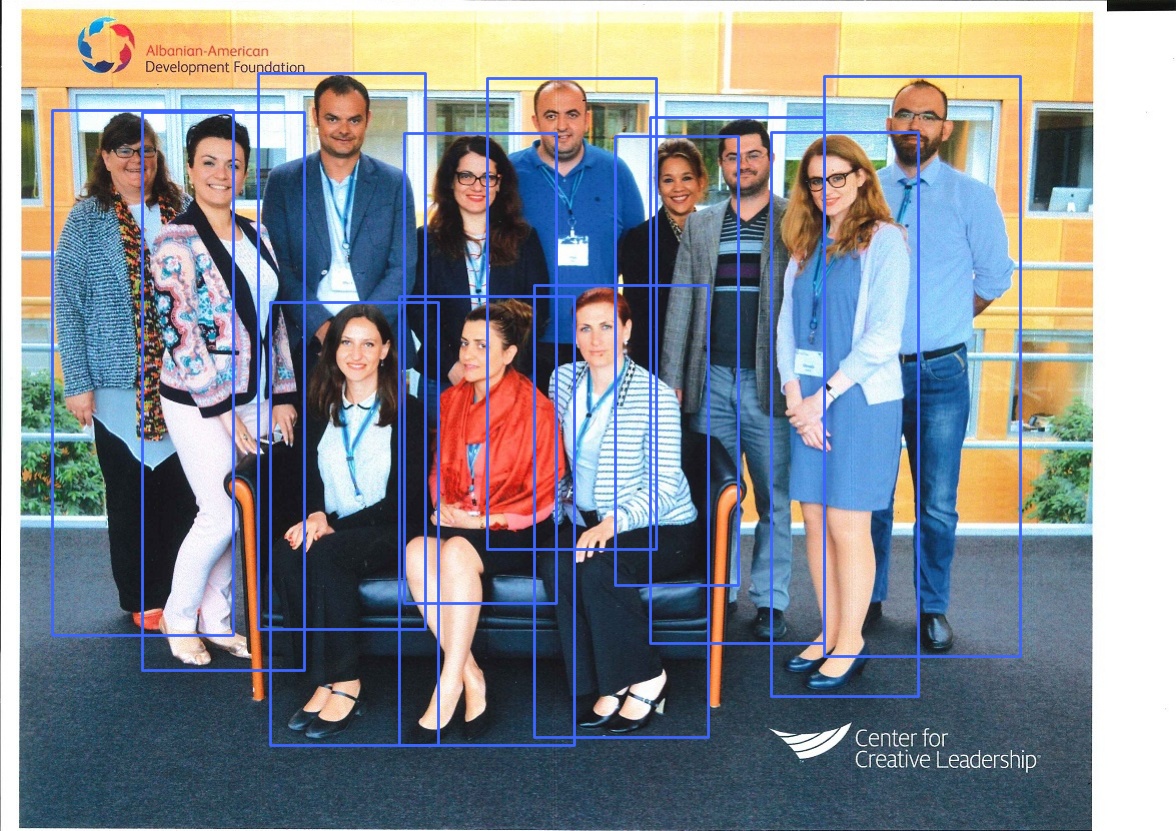}
	\includegraphics[width = .3\linewidth]{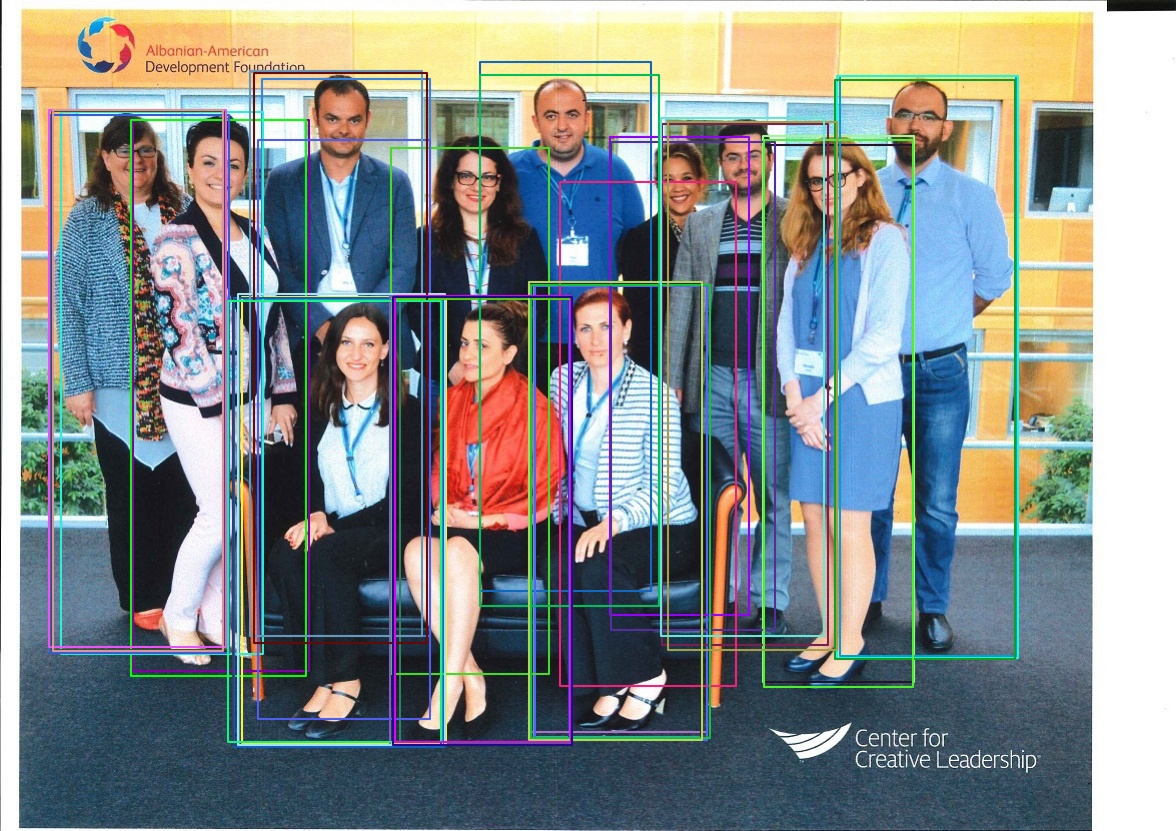}
	\includegraphics[width = .3\linewidth]{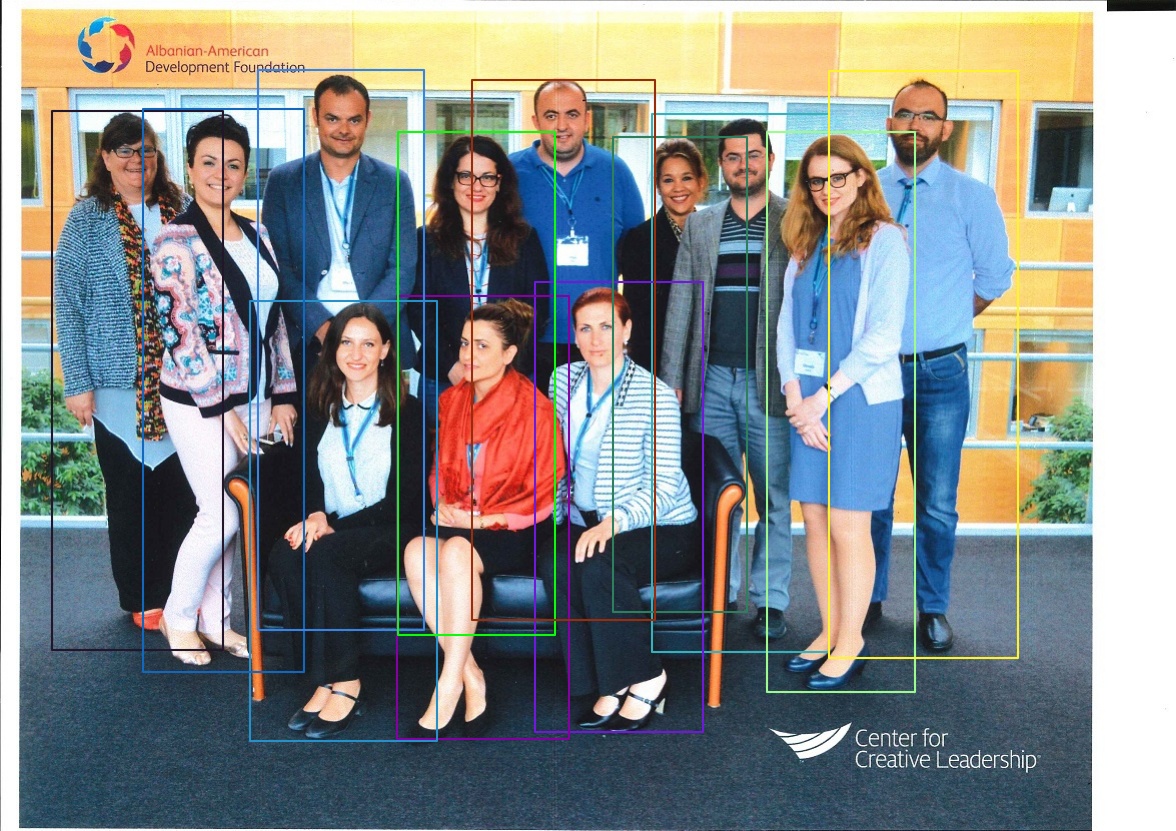}
	
	\subcaptionbox{Ground-truth}{\includegraphics[width = .3\linewidth]{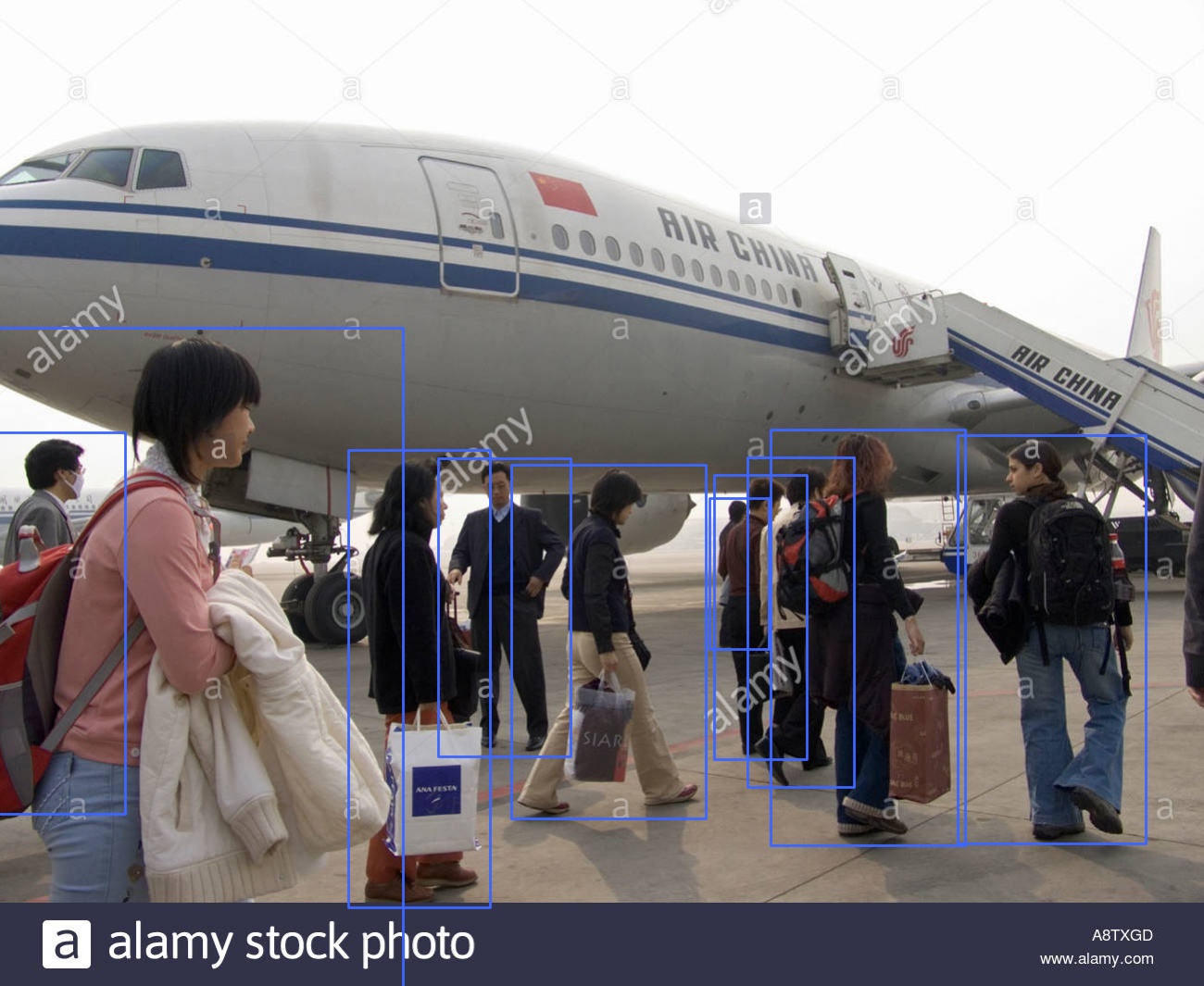}}
	\subcaptionbox{FCOS w/o NMS}{\includegraphics[width = .3\linewidth]{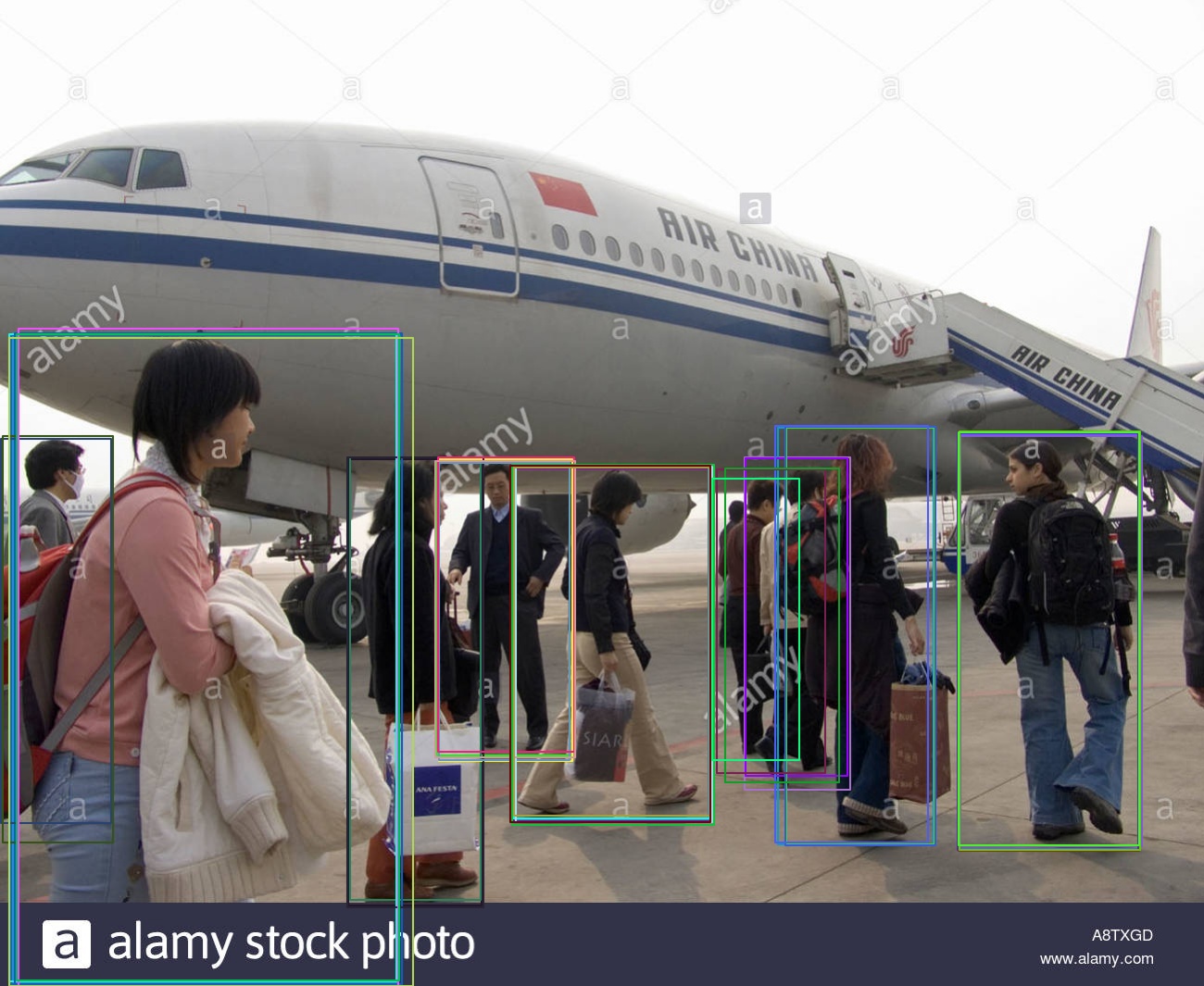}}
	\subcaptionbox{Ours}{\includegraphics[width = .3\linewidth]{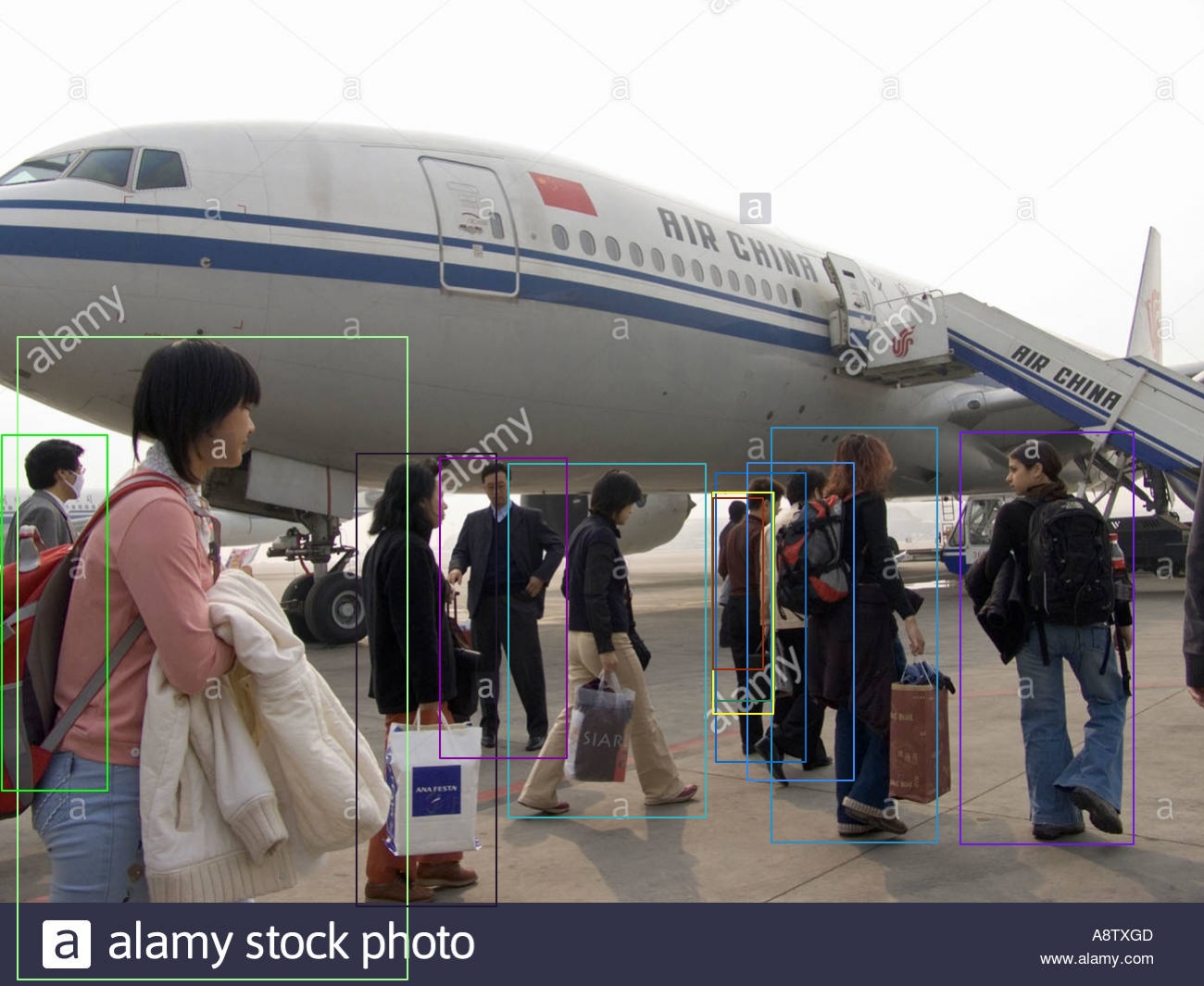}}
	
	\caption{The prediction visualizations of different detectors on CrowdHuman val set.}
	\label{fig9}
\end{figure}

Fig.~\ref{fig9} presents the detection results on the CrowdHuman dataset. We can observe similar phenomenons that FCOS suffers from severe duplicated predictions without NMS, while our end-to-end
detector obtains much fewer duplicated predictions.

{\small
\bibliographystyle{ieee_fullname}
\bibliography{reference}
}

\end{document}